\RequirePackage[2020-02-02]{latexrelease}
\documentclass{nle}
\usepackage[T1]{fontenc}
\usepackage{amsmath}
\usepackage{amsfonts}
\usepackage{graphicx}
\usepackage{natbib}
\ifpdf%
\usepackage{epstopdf}%
\else%
\fi
\usepackage{multirow}
\usepackage[inline]{enumitem}
\usepackage{pgffor}
\usepackage{url}

\usepackage{xcolor}
\definecolor{blue}{rgb}{0.1216, 0.4706, 0.7059}
\definecolor{red}{rgb}{0.8902, 0.102, 0.1098}
\definecolor{egreen}{RGB}{30,153,67}

\definecolor{cb_green2}{RGB}{102,194,165}
\definecolor{cb_red}{RGB}{252,141,98}
\definecolor{cb_blue}{RGB}{141,160,203}
\definecolor{cb_purple}{RGB}{231,138,195}
\definecolor{cb_green1}{RGB}{166,216,84}
\definecolor{cb_yellow}{RGB}{255,217,47}
\definecolor{cb_salmon}{RGB}{229,196,148}
\definecolor{cb_gray}{RGB}{179,179,179}

\definecolor{paired1}{RGB}{166,206,227}
\definecolor{paired2}{RGB}{31,120,180}
\definecolor{paired3}{RGB}{178,223,138}
\definecolor{paired4}{RGB}{51,160,44}
\definecolor{paired5}{RGB}{251,154,153}
\definecolor{paired6}{RGB}{227,26,28}
\definecolor{paired7}{RGB}{253,191,111}
\definecolor{paired8}{RGB}{255,127,0}
\definecolor{paired9}{RGB}{202,178,214}
\definecolor{paired10}{RGB}{106,61,154}
\definecolor{paired11}{RGB}{255,255,153}
\definecolor{paired12}{RGB}{177,89,40}

\usepackage{tikz}
\usetikzlibrary{matrix,positioning,shapes.geometric, arrows, fit}
\usetikzlibrary{calc}

\tikzstyle{module_color} = [fill=paired1!90]
\tikzstyle{sequence_color} = [fill=paired11!50]
\tikzstyle{target_sequence_color} = [fill=paired7!100]
\tikzstyle{output_color} = [fill=paired3!80]
\tikzstyle{bert_output_color} = [sequence_color]

\tikzstyle{module} = [
    rectangle, rounded corners, text centered,
    module_color,
    minimum height=.8cm, draw=black
]
\tikzstyle{large_module} = [
    module,
    text width=7cm,
    minimum width=7cm,
]

\tikzstyle{bert_module} = [
    module,
    text width=9cm,
    minimum width=10cm,
]

\tikzstyle{small_module} = [
    module,
    text width=1.5cm,
    minimum width=1.5cm,
]

\tikzstyle{sequence} = [
    minimum height=0.6cm,
    rectangle, text centered,
    draw=black, sequence_color,
]
\tikzstyle{target_sequence} = [
    sequence, target_sequence_color,
]
\tikzstyle{output} = [
    rectangle, text centered,
    draw=black, output_color
]

\tikzstyle{arrow} = [thick,->,>=stealth]
\tikzset{
    *|/.style={
        to path={
            (
            perpendicular cs: 
            horizontal line through={(\tikztostart)},
            vertical line through={(\tikztotarget)}
            )
            -- (\tikztotarget) \tikztonodes
        }
    }
}

\tikzset{
    box/.style={rectangle,draw=black,thick, minimum size=0.01cm},
}

\def\rowdist{1.5}
\def\matrixheight{0.8}
\def\matrixwidth{0.4}

\newcommand*{\drawbertoutput}[2]{%
    \begin{scope}[shift={(#1-\matrixwidth/2, #2-\matrixheight/2)}]
        \foreach \x in {0,0.1,...,0.4}{
            \foreach \y in {0,0.1,...,0.8}{
                \draw [line width=0.05pt, bert_output_color, draw=black] (\x, \y) rectangle (\x+0.1, \y+0.1);
            }%
        }%
    \end{scope}
}

\newcommand*{\drawberttargetoutput}[2]{%
    \begin{scope}[shift={(#1-\matrixwidth/2, #2-\matrixheight/2)}]
        \foreach \x in {0,0.1,...,0.4}{
            \foreach \y in {0,0.1,...,0.8}{
                \draw [target_sequence_color, line width=0.05pt, draw=black] (\x, \y) rectangle (\x+0.1, \y+0.1);
            }%
        }%
    \end{scope}
}

\usepackage{xspace}
\newcommand{\langno}{42\xspace}
\newcommand{\familyno}{10\xspace}

\newcommand{\taskno}{247\xspace}
\newcommand{\postagcombination}{14\xspace}

\newcommand{\bertmask}{\textcolor{blue}{[M]}\xspace}
\newcommand{\probetarget}[1]{{\textcolor{red}{\bf\xspace#1}}}
\newcommand{\pertname}[1]{\textsc{#1}\xspace}
\newcommand{\task}[3]{{$\langle$#1, #2, #3$\rangle$\xspace}}
\newcommand{\bitask}[2]{{$\langle$#1, #2$\rangle$\xspace}}
\newcommand{\mbert}{mBERT\xspace}
\newcommand{\xlm}{XLM-RoBERTa\xspace}
\newcommand{\wordpiece}{wordpiece\xspace}
\newcommand{\wordpieces}{wordpieces\xspace}

\newcommand{\pBtwo}{\textsc{b$_{2}$}\xspace}

\newcommand{\pLtwo}{\textsc{l$_{2}$}\xspace}

\newcommand{\pRtwo}{\textsc{r$_{2}$}\xspace}
\newcommand{\ppermute}{\textsc{permute}\xspace}
\newcommand{\pTARG}{\textsc{targ}\xspace}

\newcommand{\appref}[1]{Appendix~\ref{app:#1}}
\newcommand{\tabref}[1]{Table~\ref{tab:#1}}
\newcommand{\secref}[1]{Section~\ref{sec:#1}}
\newcommand{\figref}[1]{Figure~\ref{fig:#1}}
\newcommand{\equref}[1]{Equation~\ref{eq:#1}}

\newif{\ifhidecomments}
\hidecommentstrue

\ifhidecomments
    \newcommand{\nascomment}[1]{}
    \newcommand{\kornai}[1]{}
    \newcommand{\roy}[1]{}
    \newcommand{\judit}[1]{}
    \newcommand{\hendre}[1]{}
\else
    \newcommand{\nascomment}[1]{\textcolor{red}{[#1 ({\bf Noah})]}}
    \newcommand{\kornai}[1]{\textcolor{blue}{[#1 ({\bf Andras})]}} 
    \newcommand{\roy}[1]{\textcolor{orange}{[#1 ({\bf Roy})]}} 
    \newcommand{\judit}[1]{\textcolor{brown}{[#1 ({\bf Judit})]}} 
    \newcommand{\hendre}[1]{\textcolor{egreen}{[#1 ({\bf Endre})]}}
\fi
\newif{\ifhidetodo}
\hidetodotrue

\ifhidetodo
    \newcommand{\todo}[1]{}
\else
    \newcommand{\todo}[1]{\textcolor{magenta}{[TODO: #1 ]}}
\fi
\newcommand{\com}[1]{}
\newcommand{\camready}[1]{}
\newcommand{\resolved}[1]{}

\begin{document}
\label{firstpage}

\newcommand{\ourtitle}[0]{Morphosyntactic Probing of Multilingual BERT Models}

\lefttitle{\ourtitle}
\righttitle{Natural Language Engineering}

\papertitle{Article}

\jnlPage{1}{00}
\jnlDoiYr{2022}
\doival{10.1017/xxxxx}

\title{\ourtitle}

\begin{authgrp}
    \author{Judit Acs$^{1,2}$}
    \author{Endre Hamerlik$^{1,3}$}
    \author{Roy Schwartz$^4$}
   \author{Noah A.~Smith$^{5,6}$}
    \author{ Andras Kornai$^{1,2}$}

  \affiliation{$^1$ \textbf{ELKH Institute for Computer Science and Control (SZTAKI)}\\
 Budapest,  Hungary \\
\textit{
  	 \email{acsjudit@sztaki.hu},
	\email{kornai@sztaki.hu},
	\email{hamerlik.endre@sztaki.hu}
	}
}
  \affiliation{$^2$\textbf{Budapest University of Technology and Economics}  \\
Budapest, Hungary\\
\textit{
    	\email{kornai@math.bme.hu,}
	\email{judit@aut.bme.hu}
	} 
}
  \affiliation{$^3$\textbf{Department of Applied Informatics, Comenius University in Bratislava Faculty of Mathematics Physics and Informatics}  \\
Bratislava, Slovakia\\
\textit{
    	\email{endre.hamerlik@fmph.uniba.sk}
	} 
}
  \affiliation{$^4$\textbf{Hebrew University of Jerusalem} \\
Jerusalem, Israel \\
\textit{
  	\email{roy.schwartz1@mail.huji.ac.il}
	}
}
  \affiliation{$^5$\textbf{Paul G.~Allen School of Computer Science
      and Engineering, University of Washington} \\
Seattle, WA, USA  \\
\textit{
    	\email{nasmith@cs.washington.edu}
	}
}
  \affiliation{$^6$\textbf{Allen
      Institute for Artificial Intelligence}\\
Seattle, WA, USA  \\
\textit{
    	\email{noah@allenai.org}
	}
}

  \end{authgrp}

\begin{abstract}

We introduce an extensive dataset for multilingual probing of
morphological information in language models (\taskno tasks across \langno
languages from \familyno families), each consisting of a sentence with a target
word and a morphological tag as the desired label, derived from the Universal
Dependencies treebanks.  We find that pre-trained Transformer models (\mbert
and \xlm) learn features that attain strong performance across these tasks.  We
then apply two methods to locate, for each probing task, where the
disambiguating information resides in the input.  The first is a new
perturbation method that ``masks'' various parts of context; the second is the
classical method of Shapley values.  The most intriguing finding that emerges
is a strong tendency for the preceding context to hold more information
relevant to the prediction than the following context.

\end{abstract}

\maketitle

\section{Introduction}\label{sec:introduction}

The latest generation of masked language models (MLMs), which have demonstrated
great success in practical applications, have also been the object of direct
study \citep{Conneau:2018b, Liu:2019, Tenney:2019, Warstadt:2019,
Belinkov:2017, Bisazza:2018, Belinkov:2022, Ravichander:2021}. To what extent
do these models play the role of grammarians, rediscovering and encoding
linguistic structures like those found in theories of natural language syntax?
In this paper, our focus is on \textbf{morphology}; since morphological systems
vary greatly across languages, we turn to the \textbf{multilingual} variants of
such models, exemplified by \mbert \citep{Devlin:2019} and \xlm
\citep{Conneau:2020b}.

We first introduce a new morphological probing dataset of \taskno probes,
covering \langno languages from \familyno families (Section~\ref{sec:data})
sampled from the Universal Dependencies Treebank
\citep{Nivre:2020}.\footnote{Dataset, code, and full results are available at
\url{https://github.com/juditacs/morphology-probes}.}  As we argue in
Section~\ref{sec:related}, this new dataset, which includes ambiguous word
forms in context, enables substantially more extensive explorations than those
considered in the past. To the best of our knowledge, this is the most
extensive multilingual morphosyntactic probing dataset.

Our second contribution is an extensive probing study
(Sections~\ref{sec:methods}--\ref{sec:ablations}), focusing on \mbert and \xlm.
We find that the features they learn are quite strong, outperforming an LSTM
that treats sentences as sequences of characters, but which does not have the
benefit of language model pre-training. Among other findings, we observe that
\xlm's larger vocabulary and embedding are better suited for a multilingual
context than \mbert's, and, extending the work of \citet{Zhang:2018} on
recurrent networks, Transformer-based MLMs may memorize word identities and
their configurations in the training data. Our study includes several ablations
(Section~\ref{sec:ablations}) designed to address potential shortcomings of
probing studies raised by \cite{Belinkov:2022} and \cite{Ravichander:2021}.

Finally, we aim to shed light not only on the models, but on how linguistic
context cues morphological categorization.  Specifically, \emph{where} in the
context does the information reside?  Because our dataset offers a large number
(\taskno) of tasks, emergent patterns may correspond to general properties of
language.  Our first method (Section~\ref{sec:pert}) perturbs probe task
instances (both at training time and test time).  Perturbations include masking
the probe instance's target word, words in the left and/or right context, and
permuting the words in a sentence. Unsurprisingly, the target word itself is
most important.  We measure the effect on the probe's accuracy and find that
patterns of these effects across different perturbations tend to be similar
within typological language groups. 

The second method (Section~\ref{sec:shapley}) builds on the notion of
perturbations and seeks to assign responsibility to different positions in the
context of a word, using \citet{Shapley:1951} values.  We find a tendency
across most tasks to rely more strongly on \emph{left} context than on right
context.\footnote{For languages with right-to-left orthography, we reverse
ordering so that ``left'' always means ``earlier in the sequence'' and
``right'' always means ``later in the sequence.''}  Given that there is no
directional bias in Transformer-based models like \mbert and \xlm, this
asymmetry---that morphological information appears to spread
progressively---is quite surprising but significant.\footnote{It is significant at the 95\% confidence
level. It holds true in 172 of the \taskno tasks, $p$-value for binomial sign
test $6.26\cdot10^{-5}$.}  Moreover, the few cases where it does not hold have
straightforward linguistic explanations.

Though there are limitations to this study (e.g., the \langno languages we
consider are dominated by Indo-European languages), we believe it exemplifies
a new direction in corpus-based study of phenomena across languages, through
the lens of language modeling, in combination with longstanding annotation and
analysis methods (e.g., Shapley values). Remarkably, we can generally tie the
exceptions to the dominant Shapley pattern to language-specific typological
facts (see \ref{ss:outliers}), which goes a long way towards explaining the
reasonable (though imperfect) recovery (see Section~\ref{sec:pert}) of the
standard linguistic typology based on perturbation effects alone.

\section{Related Work and Background}\label{sec:related}

The observation that morphosyntactic features can simultaneously impact more
than one word goes back to antiquity: Apollonius Dyscolus in
the Greek, and P\={a}\d{n}ini in the Indian tradition both explained 
the phenomenon by agreement rules that are typically not at all sensitive to
linear order \citep{Householder:1981,Kiparsky:2009}. 
This is especially clear for the Greek and Sanskrit cases, where the word
order is sufficiently free for the trigger to come sometimes before, and
sometimes after, the affected (target) word.\footnote{In fact, the literature often uses
`direction' in a different sense, that of control: ``We call the element which
determines the agreement (say the subject noun phrase) the `controller'. The
element whose form is determined by agreement is the `target'. The syntactic
environment in which agreement occurs is the `domain' of agreement. And when
we indicate in what respect there is agreement (agreement in number, e.g.), we
are referring to `agreement features'. As these terms suggest, there is a
clear intuition that agreement is directional.'' \citep{Corbett:1998}}

The direction of control can be sensitive to the linking category as well
\citep{Deal:2015}, but in this paper we will speak of directionality only in
terms of temporal `before-after' order, using `left context' to mean words
preceding the target and `right context' for words following
it. Also, we use
`target' only to mean the element probed, irrespective of whether it is
controlling or controlled (a decision not always easy to make).

Qualitative micro-analysis of specific cases has been performed on many
languages with diametrically different grammars
\citep{Lapointe:1992,Lapointe:1990,Brown:2001,Adelaar:2005,Anderson:2005,Anderson:2006},
but quantitative analyses supported by larger datasets are largely absent \citep{Can:2022}.
Even if so, the models in question are attention-free ones which are shown to be 
insufficient to deal with long-term dependencies \citep{Li:2020}.

Here we take advantage of the recent appearance of both data and models
suitable for large-scale quantitative analysis of the directional spreading of
morphosyntactic features. The data, coming from Universal Dependencies (UD)
treebanks (see \ref{ssec:data_generation}) has suitable per-token but
contextual representations for full sentences or paragraphs.

The models take advantage of the recent shift from the standard,
identity-based, categorical variable-like unary treatment of the words. This
shift was, perhaps, the main factor contributing to the success of neural
network based language modeling. The internal representations of such models, 
often manifested in their hidden activations,
proved to be a fruitful encoding \citep{Bengio:2000}. These {\it word
  embeddings}, as they came to be known, are low-dimensional numerical
representations, typically real-valued vectors.  Thanks to their ability to
solve semantic and linguistic analogies both Word2Vec \citep{Mikolov:2013} and
GloVe \citep{Pennington:2014} gathered great interest. For a recent granular
survey of progress towards pre-trained language models see
\citet{Qiu:2020} and 
\citet{Belinkov:2019}, and for reviewing the literature on probing internal
representations see \citet{Belinkov:2022}.

\subsection{Contextual Language Models}
\label{sec:models}

Contextual language models took the relevance of the context further in that
they take a long sequence of words (even multiple sentences) as their input and
assign a vector to each segment, typically a subword, so that the same word has
distinct representations depending on its context.  One of the first widely
available contextual models was ELMo \citep{Peters:2018}, which handled
homonymy and polysemy much better than the context-independent embeddings,
resulting in a significant performance increase on downstream NLP tasks when
used in combination with other neural text classifiers
\citep{Peters:2018,Qiu:2020}. Another major improvement in line was the
introduction of Transformer-based \citep{Vaswani:2017} masked language models,
and their embeddings.

\subsubsection{Multilingual BERT}

BERT is a language model built on Transformer layers. \citet{Devlin:2019}
introduced two BERT `sizes', a base model and a large model. BERT-base has 12
Transformer layers with 12 attention heads. The hidden
size of each layer is 768. BERT-large has 24 layers with 16 heads and 1024
hidden units.  BERT-base has 110M/86M parameters with/without the embeddings, BERT-large
has 340M/303M parameters with/without the embeddings. The size of the embedding depends
on the size of the vocabulary which is specific to each pre-trained BERT model.

Multilingual BERT (mBERT) was released along with BERT, supporting 104
languages.  The main difference is that mBERT is trained on text from many
languages.  In particular, it was trained on
resource-balanced\footnote{Languages with many Wikipedia articles were
undersampled while the low-resource languages oversampled.} Wikipedia dumps
with a shared vocabulary across the supported languages.  As a BERT-base model,
its 12 Transformer layers have 86M parameters, while its large vocabulary
requires an embedding with 92M additional parameters.\footnote{In comparison, the English
    BERT-base model has a much smaller vocabulary and therefore a smaller
    embedding with 23M parameters. The combined parameter count of the English
BERT, 110M parameters, is mistakenly listed as the parameter count of \mbert on
the website
(\url{https://github.com/google-research/bert/blob/master/multilingual.md}).}

\subsubsection{\xlm}

XLM-RoBERTa is a hybrid model mixing together features of two popular 
Transformer-based models, XLM \citep{Lample:2019} and RoBERTa \citep{Liu:2019a}.

XLM is trained on both masked language modeling (MLM) and Translation Language
Modeling (TLM) objective on parallel sentences. In contrast, XLM-RoBERTa is
trained using the MLM objective only, like RoBERTa. The main difference between
XLM-RoBERTa and RoBERTa remains the scale of the corpora they were trained on:
XLM-RoBERTa's multilingual training corpora counts five times more tokens and
more than twice as many (278M with embeddings) parameters than RoBERTa's 124M
\citep{Conneau:2020b}. Another major difference between these two models is
that XLM-RoBERTa is trained in self-supervised manner, while the parallel
corpora for XLM is a supervised teaching signal. In the Cross-lingual Natural
Language Interface (XNLI; \citealp{Conneau:2018c}) evaluation of mBERT, XLM,
and XLM-RoBERTa, the latter outperformed the other MLMs in all the languages
tested by \citet{Conneau:2020b}. 

\subsubsection{Directionality}\label{ssec:directionality}

One of the Transformer architecture's main novelties was the removal of
recurrent connections, and thereby discarding the ordering of the input
symbols. Instead of recurrent connections, word order is expressed through
positional encoding, a simple position-dependent value added to the subword
embedding. Transformers have no inherent bias towards directionality. This
means that our results on the asymmetrical nature of morphosyntax
(c.f.~\secref{discussions}) can only be attributed to the language, rather than
the model.

\subsubsection{Tokenization}\label{sec:tokenization}

The idea of an intermediate {\it subword} unit
between character and word tokenization is common to \mbert and \xlm. The
inventory of subwords is learned via simple frequency-based methods starting
with byte pair encoding (BPE; \citealp{Gage:1994}). Initially, characters are
added to the inventory, and BPE repeatedly merges the most frequent
bigrams. This process ends when the inventory reaches a predefined size. The
resulting subword inventory contains frequent character sequences, often full
words, as well as the character alphabet as a fallback when longer sequences
are not present in the input text.  During inference time, the longest
possible sequence is used starting from the beginning of the word.  \mbert
uses the WordPiece algorithm \citep{Wu:2016}, a modification of BPE.  \xlm
uses the Sentence Piece algorithm \citep{Kudo:2018}, another variant of BPE.

Each BERT model has its own vocabulary. The vocabulary is trained before the
model, not in an end-to-end fashion like the rest of the model parameters.
\mbert and \xlm both share the vocabulary across 100 languages with no
distinction between the languages. This means that a subword may be used in
multiple languages that share the same script.  The subwords are
differentiated whether they are word-initial or continuation symbols. \mbert
marks the continuation symbols by prefixing them with \verb|##|. In contrast,
\xlm marks the word-initial symbols rather then the continuation symbols, with
a Unicode lower eighth block (\verb|\u2581|). The idea is that both these marks are almost
non-existent in natural text so it is easy to recover the original token
boundaries.

\mbert uses a vocabulary with 118k subwords, while \xlm's vocabulary has 250k
subwords. This means that \xlm tends to generate fewer subwords for a given
token, because longer partial matches are found more easily. \cite{Acs:2019d}
defines a tokenizer's \emph{fertility} as the proportion of subwords to tokens.
The higher this number is, the more often the tokenizer splits a token.
\mbert's average fertility on our full probing dataset is 1.9, while \xlm's
fertility is 1.7. The target words that we probe have much higher fertility
(3.1 for \mbert and 2.6 for \xlm). We attribute this to the fact that
morphology is often expressed in affixes, making the word longer, and that
longer words tend to have more morphological labels.  Both tokenizers have the
highest fertility in Belarusian (2.6 and 2.2) out of the \langno languages we
consider in this paper. \mbert has the lowest fertility in English (1.7) and
\xlm in Urdu (1.4).

\subsection{Probing}

Learning general-purpose language representations (embeddings) is a
significant thread of the NLP research \citep{Conneau:2018a}. According to
\citet{Devlin:2019} there are two major strategies to exploit the linguistic
abilities of these internal representations of language models pre-trained
either for neural machine translation (NMT) or language modeling in
general. The \emph{feature-based} approach, such as ELMo \citep{Peters:2018},
uses dedicated model architectures for each downstream task, where pre-trained
representations are included but remain unchanged. The \emph{fine-tuning}
approach, exemplified by GPT \citep{Radford:2018}, strives to modify all the LM's
parameters with as few new task-specific parameters as possible.

From this perspective, probing is a feature-based approach, with few new
parameters.  The goal of probing is not to enrich, but rather to explain the
neural representations of the model.  Probes use auxiliary classifiers (also
called \emph{diagnostic classifiers}) hooked to a pre-trained model that has
frozen weights to train a minimal-architecture classifier -- typically linear
or an MLP -- to predict a specific linguistic feature of the input. The
performance of the classifier is considered indicative of the model's
``knowledge'' in a particular task.

Probing as an explanation method was first used to evaluate static embeddings
for part-of-speech and morphological features by \citet{Koehn:2015} and
\citet{Gupta:2015}, paving the way for other studies to extend the body of
research to semantic tasks \citep{Shi:2016, Ettinger:2016, Veldhoen:2016,
Qian:2016, Adi:2017, Belinkov:2017b, Conneau:2018a}, syntax
\citep{Hewitt:2019,Goldberg:2019, Arps:2022} and multimodal tasks as well
\citep{Karpathy:2015a,Kadar:2017}. With MLMs constantly improving the state of
the art in most of the NLP benchmark tasks \citep{Qiu:2020}, the embedding
evaluation studies turned to probe these LMs
\citep{Conneau:2018b,Liu:2019,Tenney:2019,Warstadt:2019} and contextual NMT
models \citep{Belinkov:2017,Bisazza:2018}.

Although the analysis of NMT models provided many insights by comparing NMTs'
performance in probing tasks for multiple languages, the objective to compare
the morphosyntactic features of multiple languages \citep{Koehn:2015} with the
models trained on multilingual corpora. For a wider-range of model
architectures, mostly recurrent ones, see \citet{Conneau:2018a},
\citet{Sahin:2019}, and \citet{Edmiston:2020}; for Transformer-based
architectures see \citet{Liu:2019}, \citet{Ravishankar:2019},
\citet{Reif:2019}, \citet{Chi:2020}, \citet{Mikhailov:2021} and
\citet{Shapiro:2021}.  Probing multilingual models (as opposed to NMT models)
had the advantage of not requiring huge parallel corpora for training. As a
result, most of the research community has turned to multilingual MLM probing
in recent years \citep{Ravishankar:2019, Sahin:2019, Chi:2020, Mikhailov:2021,
Shapiro:2021, Arps:2022}. Our work adds morphology to this field of NLP
engineering by: 
 
\begin{itemize}

\item Extending the number of languages included in morphological probing to 42 languages.\footnote{\citet{Sahin:2019} evaluate probes in 24, \citet{Ravishankar:2019}
in 6, \citet{Chi:2020} probes syntax in 11, \citet{Mikhailov:2021} in 4
Indo-European languages.}
\item Inclusion of \textit{ambiguous word forms}\footnote{If a specific word
  form can encode different morphological tags, \citet{Sahin:2019} and others
  filter them out.} in the probing dataset in order to make the task more
  realistic. The MLMs we probe are capable of disambiguating such word forms
  based on the context.
\item Inclusion of \textit{infrequent} words as well, as opposed to
  \citet{Sahin:2019}, whose study only considers frequent words.
\item Novel ablations and probing controls (see Section~\ref{ssec:pert}).
\item Bypassing \textit{auxiliary} pseudo-tasks such as \textit{Character bin, Tag count, SameFeat,
    Oddfeat} \citep{Sahin:2019}. Such downstream tasks target proxies (artificial features, which
  are indicative of morphological ones) rather than the actual morphological
  features we concentrate on.
\item Supporting our findings with in-depth analysis of the results by means
  of Shapley values.
\end{itemize}

\section{Data}\label{sec:data}

We define a {\it probing task} as a triple of
\task{language}{POS}{morphological feature}, following UD's naming conventions
for morphological features (tags). Each sample is a sentence with a particular
target word and a morphological feature value for it.  For example, a sample
from the task \task{English}{VERB}{Tense}, would look like ``I \emph{read}
your letter yesterday.'', where \emph{read} is the target word and \emph{Past}
is the correct tag value. 

\subsection{Choice of Languages and Tags}\label{ss:lgtag}

UD 2.9 \citep{Nivre:2020} has treebanks in 122 languages. \mbert supports 104
languages while \xlm supports 100 languages. There are 55 languages in the
intersection of these three sets. We include every language from this set except
those where it is impossible to sample enough probing data. This was
unfortunately the case for Chinese, Japanese, and Vietnamese due to the lack of
data with morphosyntactic information and in Korean due to the different tagset
used in the largest treebanks.  11 other languages have insufficient data for
sampling. In contrast with e.g. \citet{Sahin:2019}, who used UniMorph for
morphological tasks, a type level morphological dataset, UD allows studying
morphology in context (often expressed through syntax).  Moreover we extended
UD 2.9 with \citet{Kote:2019}, an Albanian treebank and with a silver standard
Hungarian dataset \citep{Nemeskey:2020}. The resulting probing dataset includes
\langno languages.

UD has over 130 different morphosyntactic tags but most of them are only used
for a couple of languages. In this work, we limit our analysis to four major
tags that are available in most of the \langno languages: Case, Gender,
Number, and Tense, and four open POS classes ADJ, NOUN, PROPN, VERB. Out of
the $4 \times 4 = 16$ POS-tag combinations, 14 are attested in our set of
languages. The missing two, \bitask{NOUN}{Tense} and \bitask{PROPN}{Tense}, are
linguistically implausible. One task, \bitask{ADJ}{Tense} is only available in
Estonian. The most common tasks are \bitask{NOUN}{Number},
\bitask{NOUN}{Gender}, and \bitask{VERB}{Number}, available in 37, 32 and 27
languages respectively.  60\% of the tasks are binary (e.g.,
\task{English}{NOUN}{Number}), 20.6\% are three-way (e.g.,
\task{German}{NOUN}{Gender}) classification problems. The rest of the tasks
have four or more classes. \task{Hungarian}{NOUN}{Case} has the most classes
with 18 distinct noun cases, followed by \task{Estonian}{NOUN}{Case},
\task{Finnish}{NOUN}{Case} and \task{Finnish}{VERB}{Case} with 15, 12, and 12
cases respectively.

\begin{figure}
    \centering
    \includegraphics[width=0.79\textwidth]{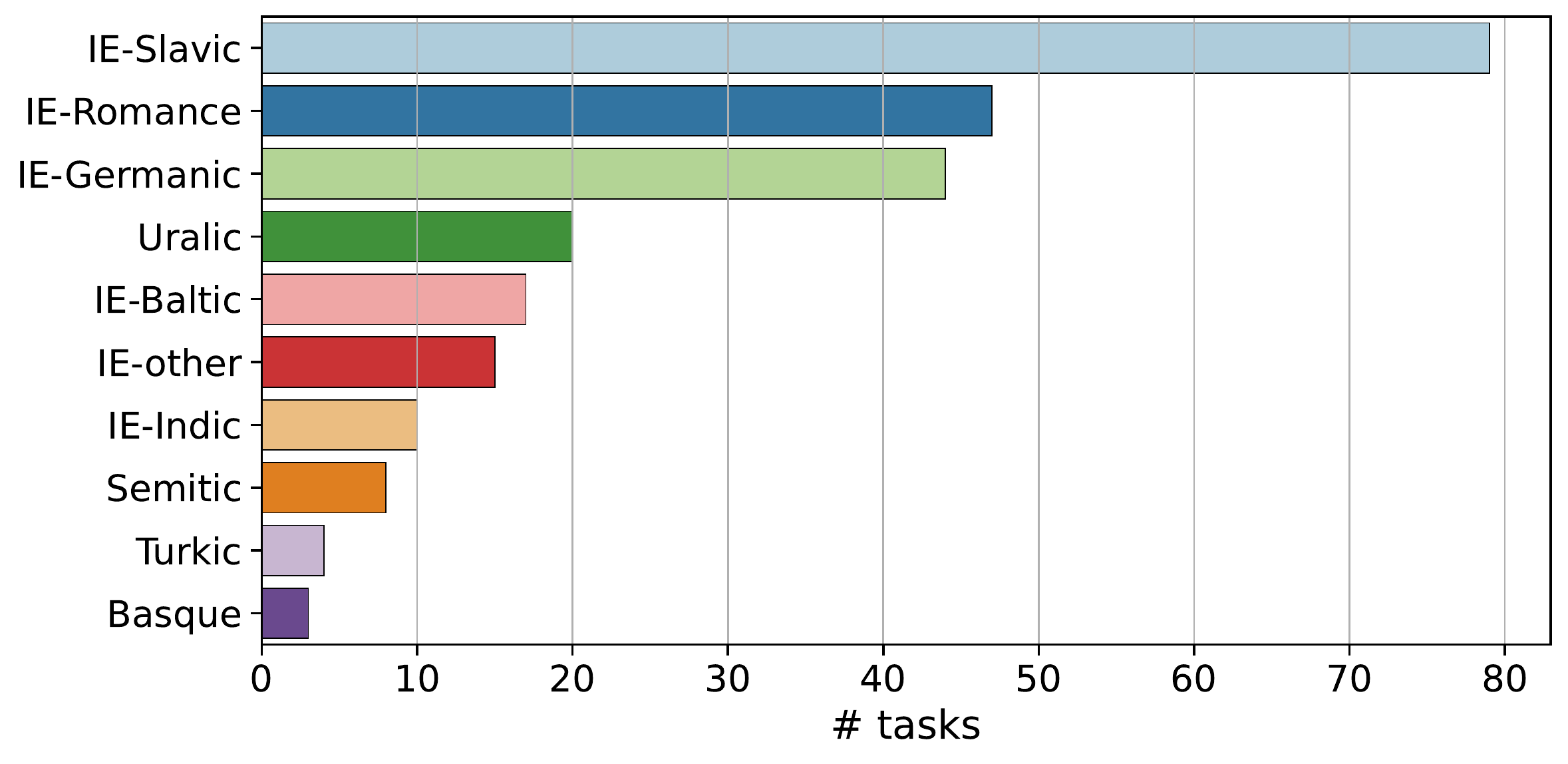}
    \vspace{1cm}
    \caption{Number of tasks by language family.}
        \label{fig:family_task_count} 
\end{figure}

\tabref{list_of_languages} lists the \langno languages included in
the probing dataset. The task counts vary greatly. We only have one task in
Afrikaans, Armenian, and Persian, while we sample 13 tasks in Russian and
12 in Icelandic.\footnote{The Icelandic UD was substantially expanded recently
\citep{Arnardottir:2020}. We were not able to sample enough data from the
earlier versions.} The resulting dataset of \taskno tasks is highly skewed
toward European languages as evidenced by \figref{family_task_count}. The
Slavic family in particular accounts for almost one third of the full dataset.
This is due to two facts. First, Slavic languages have rich morphology so most
POS/tag combinations exist in them (unlike, e.g., the Uralic languages which lack
gender). Second, there are many Slavic languages and their treebanks are very
large, the Czech treebanks are over 2M tokens, while the Russian treebanks have
1.8M tokens.  The modest number of non-European tasks is an important
limitation of our study. Fortunately the Indo-European language family is
large and diverse enough that we have examples for many different
morphosyntactic phenomena. 

\begin{table}
    \centering
    \caption{List of languages and the number of tasks in each
    language.}
    {\tablefont
\begin{tabular}{llrcllr}
\hline
     Family &          Language &  \# Tasks & $\left.\right|$ &   Family &   Language &  \# Tasks \\
\hline
     Basque &            Basque &        3 & $\left.\right|$ & IE-Slavic & Belarusian &        6 \\\hdashline
  IE-Baltic &           Latvian &       11 & $\left.\right|$ & IE-Slavic &  Bulgarian &        5 \\\hdashline
  IE-Baltic &        Lithuanian &        6 & $\left.\right|$ & IE-Slavic &   Croatian &        8 \\\hdashline
IE-Germanic &         Afrikaans &        1 & $\left.\right|$ & IE-Slavic &      Czech &       10 \\\hdashline
IE-Germanic &            Danish &        3 & $\left.\right|$ & IE-Slavic &     Polish &       10 \\\hdashline
IE-Germanic &             Dutch &        3 & $\left.\right|$ & IE-Slavic &    Russian &       13 \\\hdashline
IE-Germanic &           English &        2 & $\left.\right|$ & IE-Slavic &    Serbian &        7 \\\hdashline
IE-Germanic &            German &       10 & $\left.\right|$ & IE-Slavic &     Slovak &        8 \\\hdashline
IE-Germanic &         Icelandic &       12 & $\left.\right|$ & IE-Slavic &  Slovenian &        5 \\\hdashline
IE-Germanic &  Norwegian Bokmal &        4 & $\left.\right|$ & IE-Slavic &  Ukrainian &        7 \\\hdashline
IE-Germanic & Norwegian Nynorsk &        4 & $\left.\right|$ & IE-other &   Albanian &        6 \\\hdashline
IE-Germanic &           Swedish &        5 & $\left.\right|$ & IE-other &   Armenian &        1 \\\hdashline
   IE-Indic &             Hindi &        6 & $\left.\right|$ & IE-other &      Greek &        4 \\\hdashline
   IE-Indic &              Urdu &        4 & $\left.\right|$ & IE-other &      Irish &        3 \\\hdashline
 IE-Romance &           Catalan &        6 & $\left.\right|$ & IE-other &    Persian &        1 \\\hdashline
 IE-Romance &            French &        7 & $\left.\right|$ &   Semitic &     Arabic &        4 \\\hdashline
 IE-Romance &           Italian &        7 & $\left.\right|$ &  Semitic &     Hebrew &        4 \\\hdashline
 IE-Romance &             Latin &        9 & $\left.\right|$ & Turkic &  Turkish &        4 \\\hdashline
 IE-Romance &        Portuguese &        6 & $\left.\right|$ &   Uralic &   Estonian &        7 \\\hdashline
 IE-Romance &          Romanian &        6 & $\left.\right|$ &   Uralic &    Finnish &        8 \\\hdashline
 IE-Romance &           Spanish &        6 & $\left.\right|$ &   Uralic &  Hungarian &        5 \\
\hline
\end{tabular}

\textbf{}\\
    }
    \label{tab:list_of_languages} 
\end{table}

\medskip

\subsection{Data Generation}\label{ssec:data_generation}

UD treebanks use the CoNLL-U format, where one line corresponds to one token
and the token descriptors are separated by tabs. One such descriptor is the
morphosyntactic analysis of the token where the standard format looks like
this: \verb#MorphoTag1=Value1|MophoTag2=Value2#. This field may be empty but in
practice most non-punctuation tokens have multiple morphosyntactic tags. Some
treebanks do not include morphosyntactic tags or they use a different tagset;
we excluded these. To generate the probing tasks, we use all data available in sufficient quantity with UD tags. 

We merge treebanks in the same language but keep the train/development/test
splits and use them to sample our train, development, and test sets until we
obtain 2,000 training, 200 development and 200 test samples so that there is no
overlap between the target words in the resulting sets. We exclude languages
with fewer than 500 sentences. We limit sentence length to be between 3 and 40
tokens in the gold standard tokenization of UD. Of the candidate triples that
remain, we generate tasks where class imbalance is limited to 3:1. We attain
this by two operations: by downsampling large classes and by discarding small
classes that occur fewer than 200 times in all UD treebanks in a particular
language.\footnote{This results in some valid but rare tags not appearing in
the tasks. For example, Finnish has 15 noun cases, but 3 were too infrequent to
include in our \task{Finnish}{NOUN}{Case} task.} We discard tasks where these
sample counts are impossible to attain with our constraints. This leaves
\taskno tasks across \langno languages from \familyno language families.
Additional statistics are included in \appref{data_stats}.

\section{Methods}\label{sec:methods}

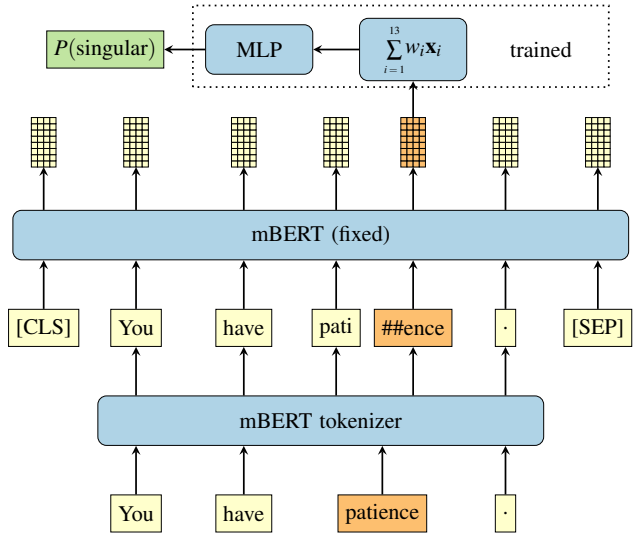
\begin{figure}
    \centering
    \resizebox{0.6\textwidth}{!}{
        \begin{tikzpicture}
            \node (tokenizer) at (0, 0) [large_module] {\mbert tokenizer};
            \node (t1) at (-3, -\rowdist) [sequence] {You};
            \node (t2) at (-1.25, -\rowdist) [sequence] {have};
            \node (t3) at (1, -\rowdist) [target_sequence] {patience};
            \node (t4) at (3, -\rowdist) [sequence] {.};
            \foreach \x in {1,...,4}
            \draw [arrow,<-] (tokenizer.south) to[*|] (t\x.north);
            \node (wp1) at (-4.5, \rowdist) [sequence] {[CLS]};
            \node (wp2) at (-3, \rowdist) [sequence] {You};
            \node (wp3) at (-1.25, \rowdist) [sequence] {have};
            \node (wp4) at (0.25, \rowdist) [sequence] {pati};
            \node (wp5) at (1.5, \rowdist) [target_sequence] {\#\#ence};
            \node (wp6) at (3, \rowdist) [sequence] {.};
            \node (wp7) at (4.5, \rowdist) [sequence] {[SEP]};
            \foreach \x in {2,...,6}
            \draw [arrow] (tokenizer.north) to[*|] (wp\x.south);
            \node[fit=(wp4)(wp5)](wp_full){};

            \node (bert) at (0, 2*\rowdist) [bert_module] {\mbert (fixed)};
            \foreach \x in {1,...,7}
            \draw [arrow,<-] (bert.south) to[*|] (wp\x.north);

            \foreach \x in {-4.5,-3,-1.25,0.25,3,4.5} \drawbertoutput{\x}{3*\rowdist} ;
            \drawberttargetoutput{1.5}{3*\rowdist} ;
            \foreach \x in {-4.5,-3,-1.25,0.25,1.5,3,4.5}
            \draw [arrow] (bert.north) to[*|] (\x, 3*\rowdist-\matrixheight/2);

            \node (weight) at (1.5, 4*\rowdist) [small_module]
                  {\shortstack{{\tiny 13}\\$\sum$\\{\tiny
                        $i=1$}}\shortstack{\phantom{n}\\ $w_i
                      \mathbf{x}_i$\\\phantom{n}}  }; 
            \node (mlp) at (-1, 4*\rowdist) [small_module] {MLP};
            \draw [arrow,<-] (weight.south) -- (1.5, 3*\rowdist+\matrixheight/2);
            \draw [arrow] (weight.west) -- (mlp.east);
            \node (output) at (-3.5, 4*\rowdist) [output] {$P(\text{singular})$};
            \draw [arrow] (mlp.west) -- (output.east);

            \draw[dotted,thick] ($(mlp.south west)+(-0.2, -0.2)$) rectangle
                ($(weight.north east)+(2.3, 0.2)$);
            \node[] at ($(weight.east)+(1.2, 0)$) {trained};

        \end{tikzpicture}
    }
    \vspace{2cm}

    \caption{Probing architecture. Input is tokenized into \wordpieces and a
        weighted sum of the \mbert layers taken on the last \wordpiece of the
        target word is used for classification by an MLP. Only the MLP
        parameters and the layer weights $w_i$ are trained. $\mathbf{x}_i$ is
        the output vector of the $i$th layer, $w_i$ is the learned layer
        weight. The example task here is \task{English}{NOUN}{Number}.
     \label{fig:tikz_architecture}}

\end{figure}

In principle, both \mbert and \xlm are trainable, but the number of parameters
is large (178M and 278M respectively), and morphologically tagged data is
simply not available in quantities that would make this feasible. We therefore
keep the models fixed, and train only a small auxiliary classifier, a multilayer
perceptron (MLP), typically with a single hidden layer and 50 neurons (for
variations see \ref{ssec:mlp_variations}) that operates on the weighted sum of
the vectors returned by each layer of the large model that is being probed.
This setup is depicted for \mbert in \figref{tikz_architecture}.

Probing as a methodology for learning about representations has had its share
of criticism \citep{Belinkov:2022,Ravichander:2021}. In particular,
\citet{Belinkov:2022} argues that probing classifiers often tell us more about
the classifier itself or the dataset than the probed model. We run several
controls and show that our results are more robust. In particular, the probing
accuracy is largely independent of the classifier hyperparameters, linear
probes are similar to non-linear probes (see
\ref{ssec:mlp_variations}); layer effects are consistent with other
probes (see \ref{ssec:layer_pooling}); fine-tuning the models is time
intensive and the results are significantly worse (see
\ref{ssec:finetuning}); and the probes work significantly better on
pre-trained checkpoints than on randomly initialized BERT models (see
\ref{ssec:randinit}).

\subsection{Baselines}\label{ssec:baselines}

Our main baseline is chLSTM, a bidirectional character\footnote{We have tried
using the subword tokenizers of both \mbert and \xlm but the results were
substantially worse and the parameter counts are very large due to the larger
embedding.} LSTM over the probing sentence. The input character sequence
(including spaces) is passed through an embedding that maps each character to a
30 dimensional continuous vector. This vector is passed along to a one-layered
LSTM with 100 hidden units. We extract the output corresponding to the first or
the last character (see \ref{ssec:subword_pooling}) and pass it to an MLP with
one hidden layer with 50 neurons (identical to the MLM probing setup). The
embedding, the LSTM and the MLP are randomly initialized and trained end-to-end
on the probing data alone. The parameter count is close to the MLM auxiliary
classifiers' parameter count (40k).
Our motivation for this model can be summarized as: 

\begin{itemize}

    \item it is contextual;

    \item it is only trained on the probing data and we can assume that if a
      MLM performs better than chLSTM, it is probably due to the MLM's
      pre-training, especially as the SIGMORPHON shared tasks are dominated by
      LSTM models;
    \item LSTMs are good at morphological inflection
        \citep{Kann:2016,Cotterell:2017}, a related but more difficult task
        than morphosyntactic classification;

    \item it is a different model family than the Transformer-based MLMs, so
        any similarity in behavior, particularly our findings using Shapley
        values explored in \secref{shapley}, is likely due to linguistic reasons rather than
        some modeling bias.

\end{itemize}

Our secondary baseline is fastText \citep{Bojanowski:2017}, a multilingual word
embedding trained on bags of character n-grams. We use the same type of MLP on
top of fastText vectors. FastText is pre-trained, though less extensively than
the MLMs.

Finally, we also run Stanza,\footnote{\tt https://stanfordnlp.github.io/stanza}
a high quality NLP toolchain for many languages. Although there are undoubtedly
better language-specific tools than Stanza for certain languages, it is outside
the scope of this paper to find the best morphosyntactic tagger for \langno
languages. The details of our Stanza setup are listed in \appref{stanza_setup}.

\subsection{Experimental Setup}

All experiments including the baselines are trained using the Adam optimizer
\citep{Kingma:2014} with $\text{lr}=0.001, \beta_1=0.9, \beta_2=0.999$. We use
early stopping based on development loss and accuracy. We use a 0.2 dropout
between the input and hidden layer of the MLP and between the hidden and the
output layers. The batch size is always set to 128 except in the fine-tuning
experiments where it is set to 8. All results throughout the paper are averaged
over 10 runs with different random seeds except the ones presented in
\secref{shapley} since they require an exponentially large number of
experiments.

\subsection{Subword Pooling}\label{ssec:subword_pooling}

FastText maps every word to a single vector, and can generate vectors for OOV
words with an off\-line script. On the other hand, \mbert and chLSTM may assign
multiple vectors to the target word. \mbert assigns a vector to each subword
and chLSTM assigns a vector to each character.  These models require a way to
pool multiple vectors that correspond to the target word.  \citet{Devlin:2019}
used the first \wordpiece of every token for named entity recognition.
\citet{Kondratyuk:2019} and \citet{Kitaev:2019} found no difference between
using first, last, or max pooling for dependency parsing and constituency
parsing in many languages. \citet{Acs:2021} showed that the last subword is
usually the best for morphology and more sophisticated pooling choices do not
improve the results, so we only compare the first and the last subword for both
\mbert and \xlm and use the better choice based on development accuracy.  This
turns out to be the last subword for 98\% of the tasks.  Similarly, we
consider the first and the last character for chLSTM. The last character is the
better choice in 82\% of the tasks.

\section{Results}\label{sec:results}

\subsection{Morphology in Pre-trained Language Models}
\label{ssec:unperturbed_results}

We first examine how well morphology can be recovered from the model
representations. Table~\ref{tab:task_model_averages} shows the average probing
accuracy on each morphological task. The average is computed over all languages
each task is available in. \xlm is slightly better than \mbert, and both are
clearly superior to chLSTM and fastText. The baselines are also close to each other
but chLSTM is 0.6\% better than fastText. Out of the \postagcombination~\bitask{POS}{tag}
combinations, \mbert is only better than \xlm in \bitask{ADJ}{Gender} but the
difference is not statistically significant ($p>0.05$ with Bonferroni
correction).\footnote{Paired $t$-test across the languages that the
\bitask{ADJ}{Gender} task is available in.} In fact \xlm is only statistically
significantly better than \mbert at 5 POS-tag combinations out of the
\postagcombination: \bitask{ADJ}{Case}, \bitask{NOUN}{Case},
\bitask{NOUN}{Number}, \bitask{VERB}{Number} and \bitask{VERB}{Tense}.  Since
chLSTM is the better baseline and it is a practical estimation of the maximum
performance achievable with the probing data alone, we limit our analysis
to chLSTM and the two masked language models.

Perhaps the most salient fact about these results is that the MLM-based systems
perform in the high 80-90\% level (only one task, \bitask{PROPN}{Gender} is at
76\%), something quite remarkable compared to the state of the art only a
decade ago \citep{Kurimo:2010b}. In fact, those current models that are tuned
to individual tasks and languages can often go beyond the performance of the
generic models presented here, but our interest is with universal morphological
claims one can distill from adapting generic MLM models to highly
language-specific tasks.\footnote{In regards to genericity the status of Stanza
is not well documented.} The auxiliary classifier has relatively few (40k)
parameters, no more than the fully task-specifically trained baselines,
nevertheless outperforms both chLSTM and fastText. This indicates clearly that
the morphological knowledge is not in the auxiliary classifier alone, some of it
must already be present in the pre-trained weights that come with \mbert and
\xlm. For detailed comparison with randomized baselines see Section
\ref{sec:ablations}.

\begin{table}
    \centering
    \caption{
        Average test accuracy over all languages by
    task and model. The last row is the average of all \taskno tasks. Stanza
does not support Albanian so the 6 Albanian tasks are not included in the
Stanza results.}
    {\tablefont
\begin{tabular}{llrrrrr}
\hline
Tag & POS &   mBERT  &   XLM-R  &   chLSTM  &   fastText  &   Stanza* \\
\hline
case & adj &   87.6 &   90.5 &    77.9 &      70.3 &     93.3 \\\hdashline
    & noun &   88.8 &   91.4 &    78.9 &      72.7 &     90.6 \\\hdashline
    & propn &   87.5 &   89.3 &    78.8 &      68.4 &     89.1 \\\hdashline
    & verb &   88.5 &   90.9 &    80.3 &      78.9 &     90.1 \\\hdashline
gender & adj &   87.6 &   86.3 &    83.4 &      80.3 &     91.9 \\\hdashline
    & noun &   87.2 &   88.6 &    78.9 &      80.9 &     90.0 \\\hdashline
    & propn &   75.6 &   76.5 &    64.2 &      69.0 &     70.4 \\\hdashline
    & verb &   88.0 &   88.5 &    87.8 &      87.0 &     86.7 \\\hdashline
number & adj &   96.4 &   97.1 &    91.7 &      89.1 &     96.3 \\\hdashline
    & noun &   93.5 &   95.1 &    88.9 &      90.1 &     94.2 \\\hdashline
    & propn &   89.2 &   90.1 &    81.2 &      78.4 &     78.7 \\\hdashline
    & verb &   95.8 &   97.0 &    95.1 &      95.2 &     95.7 \\\hdashline
tense & adj &   98.0 &   99.5 &   100.0 &      96.6 &     98.0 \\\hdashline
    & verb &   91.7 &   94.1 &    90.7 &      90.9 &     91.1 \\\hdashline
ALL &      &   90.4 &   91.8 &    85.0 &      83.3 &     91.4 \\
\hline
\end{tabular}
    }
    \label{tab:task_model_averages} 
\end{table}

\begin{figure}
    \centering
    \includegraphics[width=0.6\textwidth]{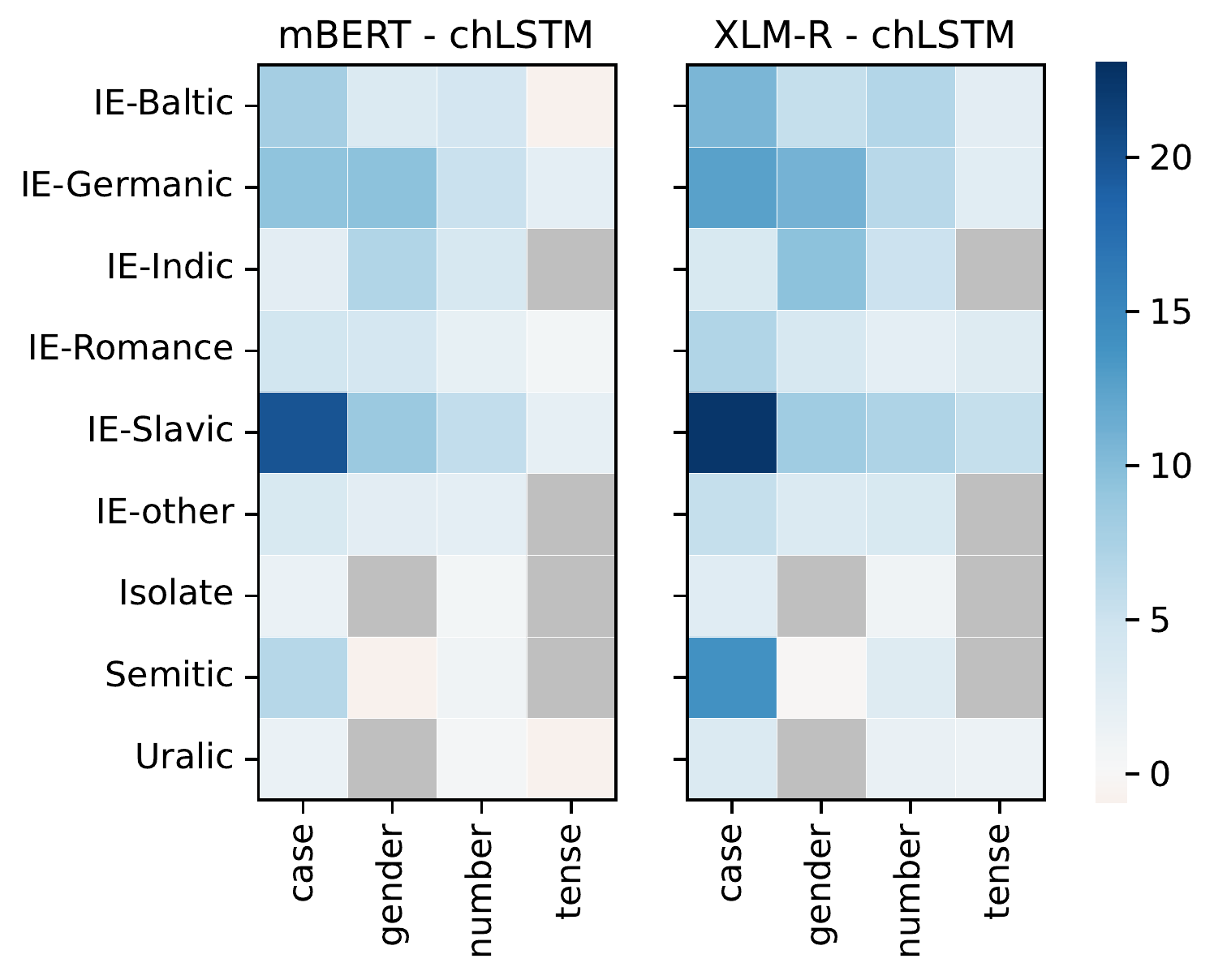}
    \vspace{1cm}
    \caption{
        Difference in accuracy between
        \mbert (left) and chLSTM, and \xlm (right) and chLSTM grouped by language
    family and morphological category. Grey cells represent missing tasks.}
        \label{fig:bert_xlm_vs_chlstm_by_family_by_tag} 
\end{figure}

\begin{figure}
    \centering
    \includegraphics[width=0.6\textwidth]{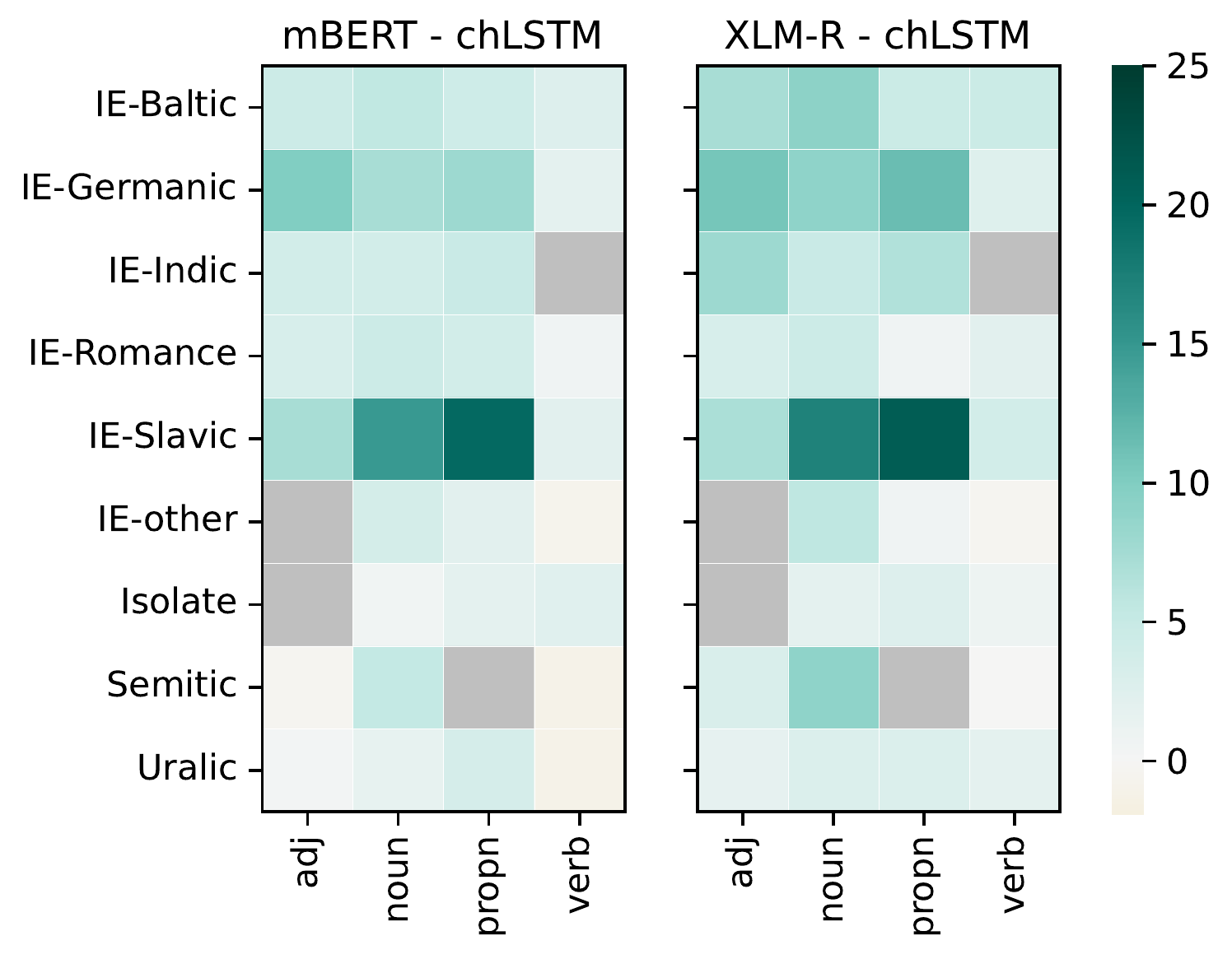}
    \vspace{1cm}
    \caption{
        Difference in accuracy between
        \mbert (left) and chLSTM, and \xlm (right) and chLSTM grouped by language
    family and POS. Grey cells represent missing tasks.}
    \label{fig:bert_xlm_vs_chlstm_by_family_by_pos} 
\end{figure}

Figures~\ref{fig:bert_xlm_vs_chlstm_by_family_by_tag} and
\ref{fig:bert_xlm_vs_chlstm_by_family_by_pos} show the difference between the
accuracy of the MLMs and chLSTM averaged over language
families.  chLSTM is only better than one or both of the pre-trained models
in 8 tasks out of the \taskno, and the difference is never large.

\begin{figure}
    \centering
    \includegraphics[width=0.7\textwidth]{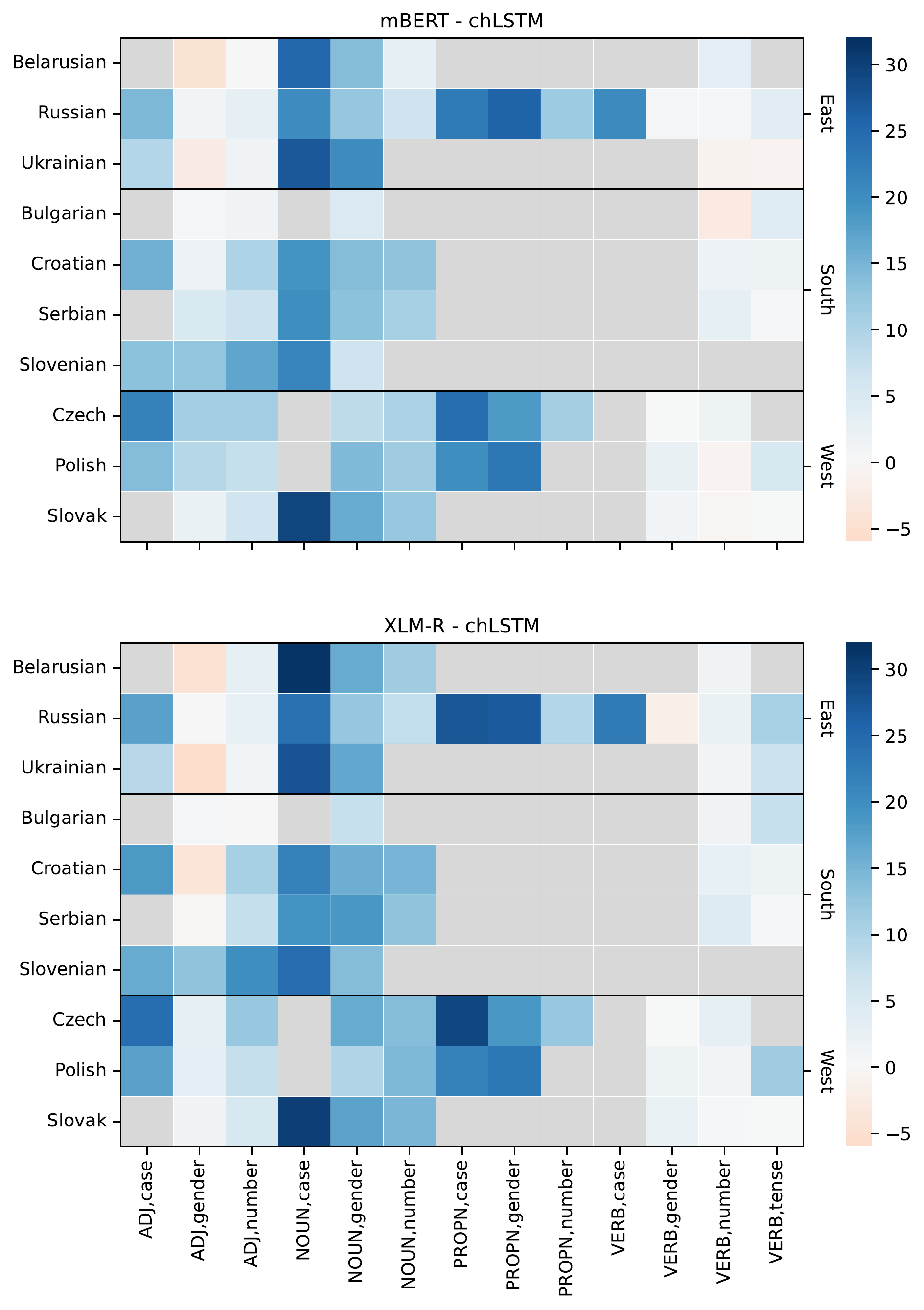}
    \vspace{1cm}
    \caption{
    Task-by-task difference between the
MLMs and chLSTM in Slavic languages. Grey cells represent
missing tasks.}
\label{fig:bert_xlm_vs_chlstm_slavic} 
\end{figure}

We find a large number of tasks at the other end of the scale. Particularly
Slavic case and gender probes work much better in both \mbert and \xlm than in
chLSTM. Slavic languages have highly complex declension with 3 genders, 6--8
cases, and frequent syncretism. This explains why chLSTM is struggling to pick
up the pattern from 2,000 training samples alone. \mbert and \xlm were both
trained on large datasets in each language and therefore may have picked up a
general representation of gender and case.\footnote{Slavic conjugation is not much
simpler, but concentrates on the word, not on the context; see
Section~\ref{sec:shapley}.} It is also worth mentioning that among the 100
languages that these models support, Slavic languages are one of the largest
language families with 10 or more languages.
\figref{bert_xlm_vs_chlstm_slavic} shows the differences for each Slavic
language and task. The similarities appear more areal (Ukrainian and Belarus,
Czech and Polish) than historical, though the major division into Eastern,
Western, and Southern Slavic is still somewhat
perceptible.

\subsection{Comparison between \mbert and \xlm}

\begin{figure}
    \centering
    \includegraphics[width=0.99\textwidth]{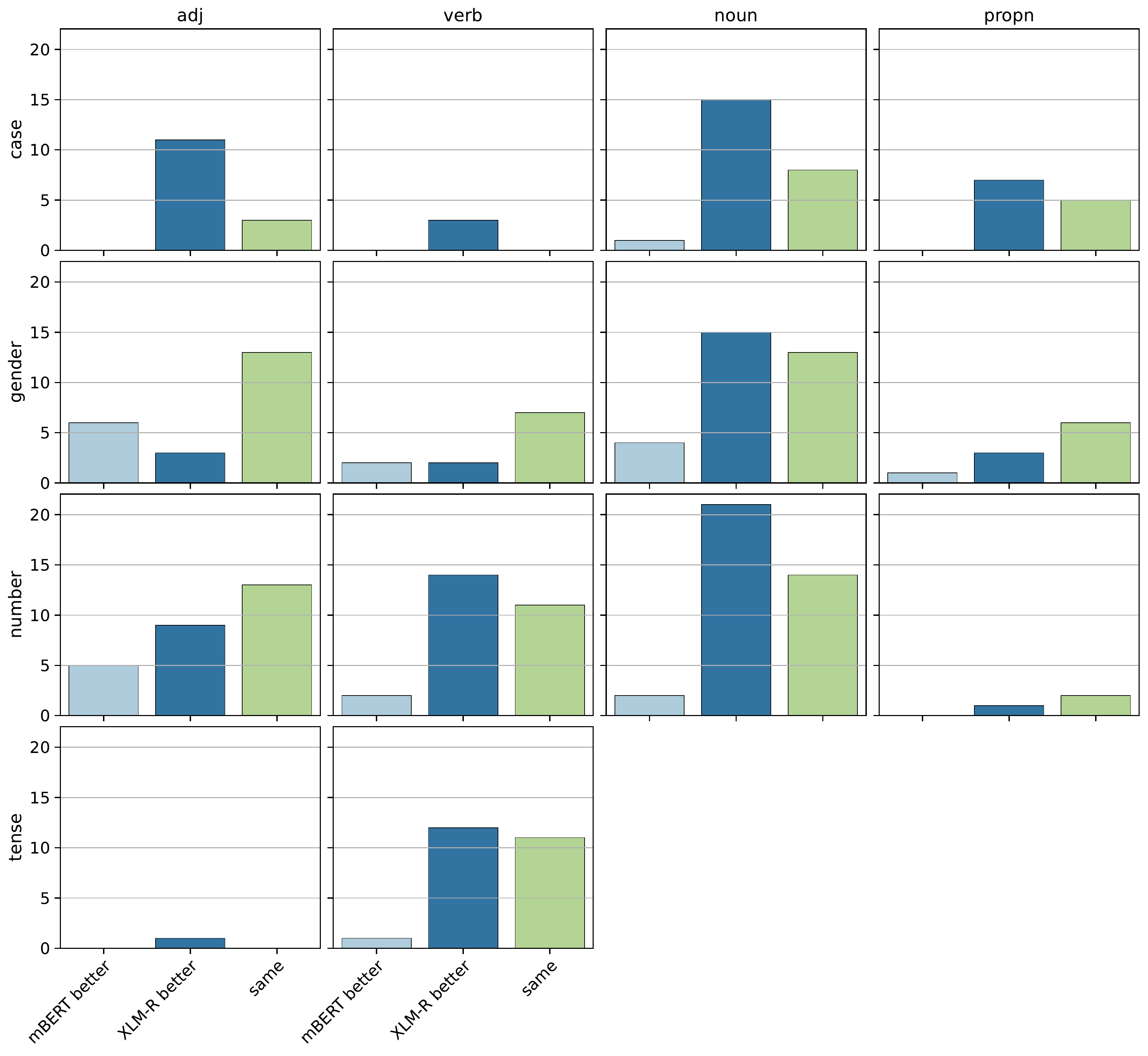}
    \vspace{1cm}
    \caption{
    \mbert \xlm comparison by tag and by POS. }
    \label{fig:bert_xlm_better}
\end{figure}

\tabref{task_model_averages} showed that \xlm is slightly better than \mbert on
average and in every POS-tag category except \bitask{ADJ}{Gender}.  However,
this advantage is not uniform over tag and POS as evidenced by
\figref{bert_xlm_better}, which shows the number of tasks where one model is
significantly better than the other. \xlm is always better or no worse than
\mbert at case and tense tasks with the exception of \task{Swedish}{NOUN}{Case}
and \task{Romanian}{VERB}{Tense}, where \mbert is the stronger model.

\begin{figure}
    \centering
    \includegraphics[width=0.99\textwidth]{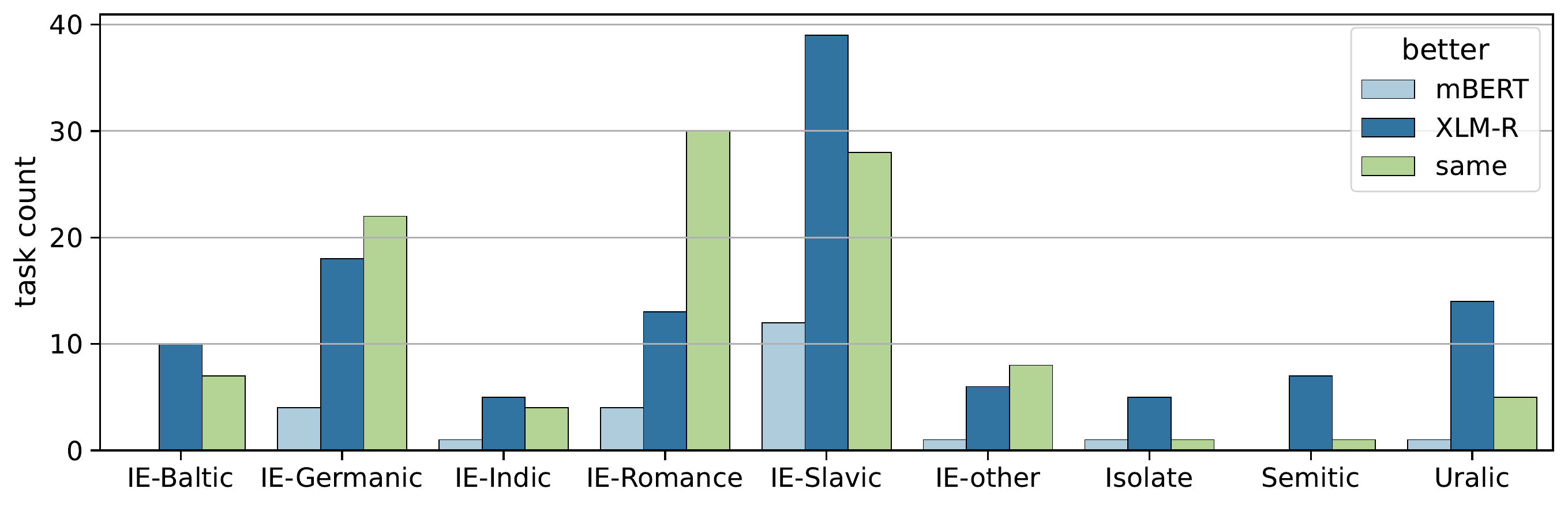}
    \vspace{1cm}
    \caption{
        \mbert \xlm comparison by language family. }
    \label{fig:bert_xlm_better_by_family}
\end{figure}

\figref{bert_xlm_better_by_family} illustrates the same task counts by language
family. We observe the same performance in most tasks from the Germanic and
Romance language families. \xlm is better at the majority of the tasks from the
Semitic, Slavic, and Uralic families, and the rest are more even.
Interestingly the two members of the Indic family in our dataset, Hindi and
Urdu behave differently. \xlm is better at 5 out of 6 Hindi tasks and the
models are the same at the sixth task. \mbert on the other hand, is better at one
Urdu task and the models are the same at other three Urdu tasks. This might be due
to the subtle differences in \mbert and \xlm subword tokenization introduced in
\ref{sec:tokenization}.

\subsection{Difficult Tasks}\label{ss:difftask}

Some morphosyntactic tags are hard to retrieve from the model representations.
In this section we examine such tags and the results in more detail.
\tabref{hard_tasks} lists the 10 hardest tasks measured by the average accuracy
of \mbert and \xlm. \task{German}{PROPN}{Case} is difficult for two reasons.
First, nouns are not inflected in German\footnote{there are some exceptions in
the genitive case}; case is marked in the article of the noun. The article
depends on both the case and the gender and syncretism (ambiguity) is very
high. This is reflected in the modest results for \task{German}{NOUN}{Case} as
well (72.9\% for \mbert, 80.7\% for \xlm). Second, proper nouns are often
multiword expressions. Since all tokens of a multiword proper noun are tagged
PROPN in UD, our sampling method may pick any of those tokens as a target token
of a probing task.

Another outlier is \task{Arabic}{ADJ}{Case}. Arabic adjectives usually follow
the noun they agree with in case. There is no agreement with the elative case, and
sometimes the adjective precedes the noun which is in genitive, but the
adjective is not. This kind of exceptionality may simply be too much to learn
based on relatively few examples -- it is still fair to say that grammarians
(humans) are better pattern recognizers than MLMs. 

\begin{table}
    \centering
    \caption{10 hardest tasks.}
    {\tablefont
\begin{tabular}{lrrrr}
\hline
                                      task &  mBERT &  XLM-R &  chLSTM & Stanza \\
\hline
$\langle$Icelandic, PROPN, gender$\rangle$ &   58.4 &   64.7 &    53.7 &   59.2 \\\hdashline
     $\langle$German, PROPN, case$\rangle$ &   66.6 &   68.8 &    64.6 &   71.5 \\\hdashline
  $\langle$Icelandic, ADJ, gender$\rangle$ &   67.5 &   70.4 &    64.7 &   82.1 \\\hdashline
       $\langle$Arabic, ADJ, case$\rangle$ &   65.8 &   72.2 &    64.2 &   81.6 \\\hdashline
   $\langle$German, PROPN, gender$\rangle$ &   70.4 &   69.1 &    53.5 &   78.6 \\\hdashline
    $\langle$Albanian, NOUN, case$\rangle$ &   69.0 &   71.8 &    67.8 &        \\\hdashline
        $\langle$Latin, ADJ, case$\rangle$ &   69.7 &   71.8 &    63.0 &   81.0 \\\hdashline
     $\langle$Irish, NOUN, gender$\rangle$ &   71.2 &   72.4 &    69.5 &   80.0 \\\hdashline
    $\langle$Dutch, PROPN, gender$\rangle$ &   70.2 &   74.8 &    62.9 &   65.0 \\\hdashline
    $\langle$Hindi, PROPN, gender$\rangle$ &   70.4 &   75.2 &    60.9 &   68.0 \\
\hline
\end{tabular}
    }
    \label{tab:hard_tasks} 
\end{table}

\section{Perturbations}\label{ssec:pert} \label{sec:pert}

In Section~\ref{ssec:perturbed_results} we analyze the MLMs' knowledge of morphology in
more detail through a set of \emph{perturbations} that remove some source of
information from the probing sentence. We
compare the different perturbations to the unperturbed MLMs, but observe
that perturbation often reduces performance to the level of the contextual
baseline (chLSTM) or even below.  The effect of major perturbations is
unmistakable. \tabref{list_of_perturbations} exemplifies each perturbation.

\begin{table}
\centering 

\caption{List of perturbation methods with examples. The target word is in
\probetarget{bold}. The mask symbol is abbreviated as \bertmask.}

{\tablefont
\begin{tabular}{lll}
\hline
Method & Explanation & Example \\
\hline
Original & & Then he ripped open Hermione 's letter and
\probetarget{read} it out loud . \\\hdashline
\pTARG & mask target word & Then he ripped open Hermione 's letter and
\probetarget{[M]} it out loud . \\\hdashline
\pLtwo & mask previous 2 words & Then he ripped open Hermione 's \bertmask \bertmask
\probetarget{read} it out loud . \\\hdashline
\pRtwo & mask next 2 words & Then he ripped open Hermione 's letter and
\probetarget{read} \bertmask \bertmask loud . \\\hdashline
\pBtwo & mask 2 on each side & Then he ripped open Hermione 's \bertmask \bertmask
\probetarget{read} \bertmask \bertmask loud . \\\hdashline
\ppermute & shuffle word order & and open \probetarget{read} Then letter . it out he ripped 's Hermione loud \\
\hline
\end{tabular}
}
\label{tab:list_of_perturbations} 
\end{table}

\paragraph{Target masking} Languages with rich inflectional morphology tend to
encode most, if not all, morphological information in the word form alone. We
test this by hiding the word form, while keeping the rest of the sentence
intact.  Recall that BERT is trained with a cloze-style language modeling
objective, i.e., 15\% tokens are replaced with a \verb|[MASK]| token and the
goal is to predict these. We employ this mask token to hide the target word
(\pertname{targ}) from the auxiliary classifier. This means that all orthographic cues
present in the word form are removed.\footnote{We use a single mask token
regardless of how many \wordpieces the target word would contain, rather than
masking each \wordpiece. Our early experiments showed negligible difference
between the two choices.}

\paragraph{Context masking} Many languages encode morphology in short phrases
that span a few words, e.g., person/number agreement  features on a verb that
is immediately preceded by a subject.  The verb tense of \textit{read}, while
ambiguous on its own, can often be disambiguated by looking at a few
surrounding words, such as the presence of an auxiliary (\textit{didn't}), or a
temporal expression.  We use the relative position of a token to the target
word, left context refers to the part of the sentence before the target word,
while right context refers to the part after it.  We try masking the left
(\pertname{l$_N$}), the right (\pertname{r$_N$}), and both sides
(\pertname{b$_N$}), where $N$ refers to the number of masked tokens.  We expand
this analysis using Shapley values in \secref{shapley}.

\paragraph{Permute} Many languages have strict constraints on the order of
words. A prime example is English, where little morphology is present at the
word level, but reordering the words can change the meaning of a sentence
dramatically.  Consider the examples \textit{Mary loves John} versus
\textit{John loves Mary}: in languages with case inflection the distinction is
made by the cases rather than the word order.  It has been shown
\citep{Sinha:2021,Ettinger:2020} that BERT models are sensitive to word order
in a variety of English and Mandarin tasks.  We quantify the importance of word
order by shuffling the words in the sentence.

\subsection{Results}
\label{ssec:perturbed_results}

\begin{table}
    \centering
    \caption{Perturbation results by model averaged over \taskno tasks. Effect is defined in \protect \equref{perturbation_effect}.}

    {\tablefont
\begin{tabular}{lrrrrrr}
\hline
perturbation &  mBERT acc &  XLM-R acc &  chLSTM acc & mBERT effect & XLM-R effect & chLSTM effect \\
\hline
 unperturbed &       90.4 &       91.8 &        85.0 &        0.00\% &        0.00\% &         0.00\% \\\hdashline
        targ &       75.8 &       80.0 &        60.2 &       16.18\% &       12.74\% &        28.90\% \\\hdashline
     permute &       84.6 &       86.1 &        83.4 &        6.74\% &        6.42\% &         3.20\% \\\hdashline
       L$_2$ &       88.4 &       90.3 &        84.2 &        2.28\% &        1.63\% &         1.10\% \\\hdashline
       R$_2$ &       89.1 &       90.9 &        85.1 &        1.48\% &        1.01\% &       --0.19\% \\\hdashline
       B$_2$ &       86.0 &       87.9 &        83.8 &        5.03\% &        4.42\% &         1.56\% \\
\hline
\end{tabular}
    }
    \label{tab:perturbation_summary} 
\end{table}

Perturbations change the input sequence or the probing setup in a way that
removes information and should result in a decrease in probing accuracy. Given
the large number of tasks and multiple perturbations, instead of listing all
individual data points, we average the results over POS, tags, and language
families, and point out the main trends and outliers. The overall average
perturbation results are listed in \tabref{perturbation_summary}.

\begin{figure}
    \centering \includegraphics[clip,trim={0 0 0
        0},width=.99\textwidth]{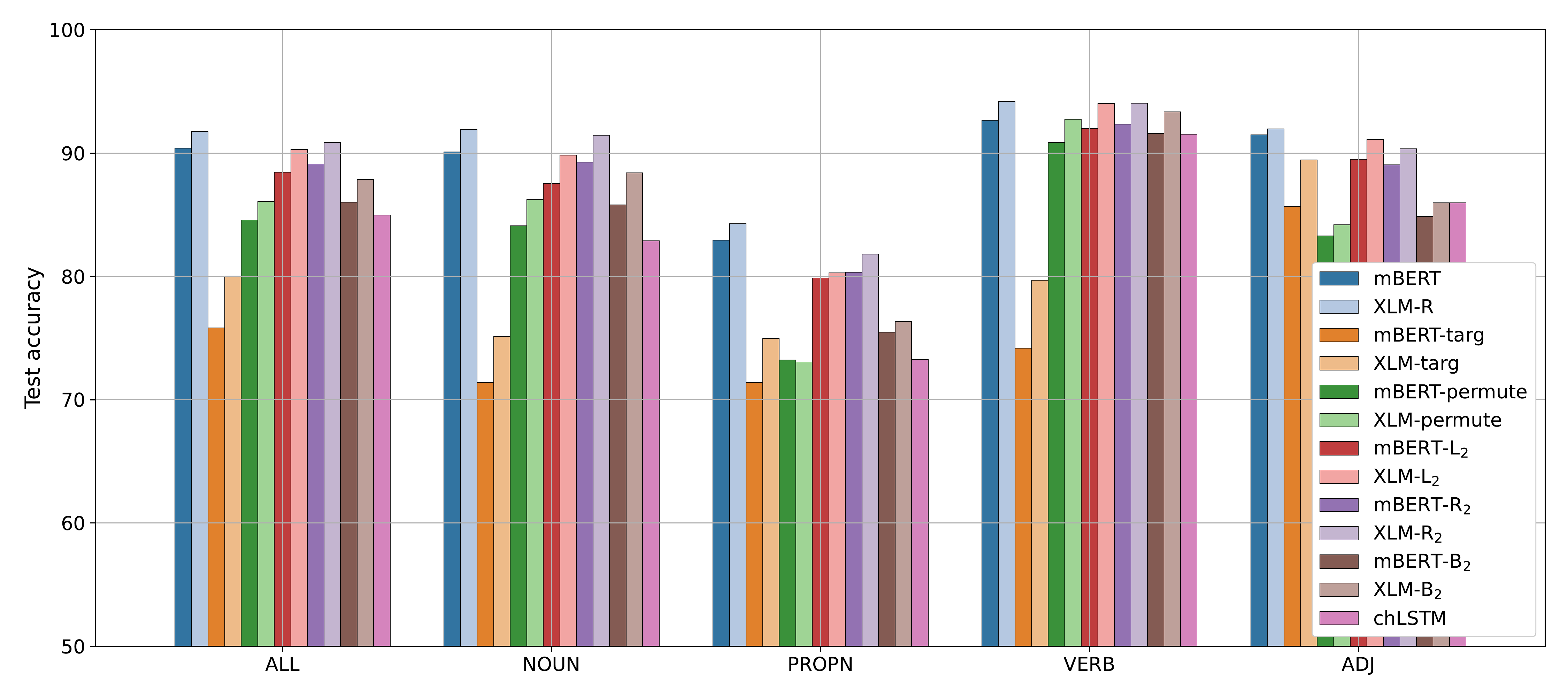}
    \vspace{1cm}

    \caption{Test accuracy of the perturbed probes grouped by POS. The first
    group is the average of all \taskno tasks. The first two bars in each group
are the unperturbed probes' accuracy.}

\label{fig:perturbation_by_pos}
\end{figure}

Our main group of perturbations involves masking one or more words in the input
sentence. Both models have dedicated mask symbols, which we use to replace
certain input words. In particular, \pTARG masks the target word, where most of
the information is contained -- precisely how much will be discussed in
Section~\ref{sec:shapley}. \ppermute shuffles the entire context, leaving the
target word fixed, \pLtwo masks the two words preceding that target word,
\pRtwo masks the two words following the target and \pBtwo masks both the
preceding two and the following two words. Remarkably, \ppermute and \pBtwo are
highly correlated, a matter we shall return to in \ref{ssec:targ_permute}.
\figref{perturbation_by_pos} shows the average test accuracy of the probes
by perturbation grouped by POS.

Since the net changes caused by masking are often quite small, particularly
for verbs, we define the {\it effect} of perturbation $p$ on task $t$ when probing
model $m$ as:

\begin{equation}
    E(m, t, p) = 1 - \frac{\text{Acc}(m, t, p)}
    {\text{Acc}(m, t)},
        \label{eq:perturbation_effect}
\end{equation}

\noindent where $\text{Acc}(m, t)$ is the unperturbed probing accuracy on task
$t$ by model $m$. We present the effect values as percentages of the original
accuracy. 50\% effect means that the probing accuracy is reduced by half.
Negative effect means that the probing accuracy \textit{improves} due to a
perturbation.

\subsubsection{Context Masking}

Proper nouns seem to be affected the most by context masking perturbations.
This is probably caused by the lack of morphological information in the word
form itself, at least in Slavic languages, where proper nouns are often
indeclinable. The models pick up much of the information from the context.  We
shall examine this in more detail in \secref{shapley}.

\begin{figure}
    \centering
    \includegraphics[clip,width=0.99\textwidth]{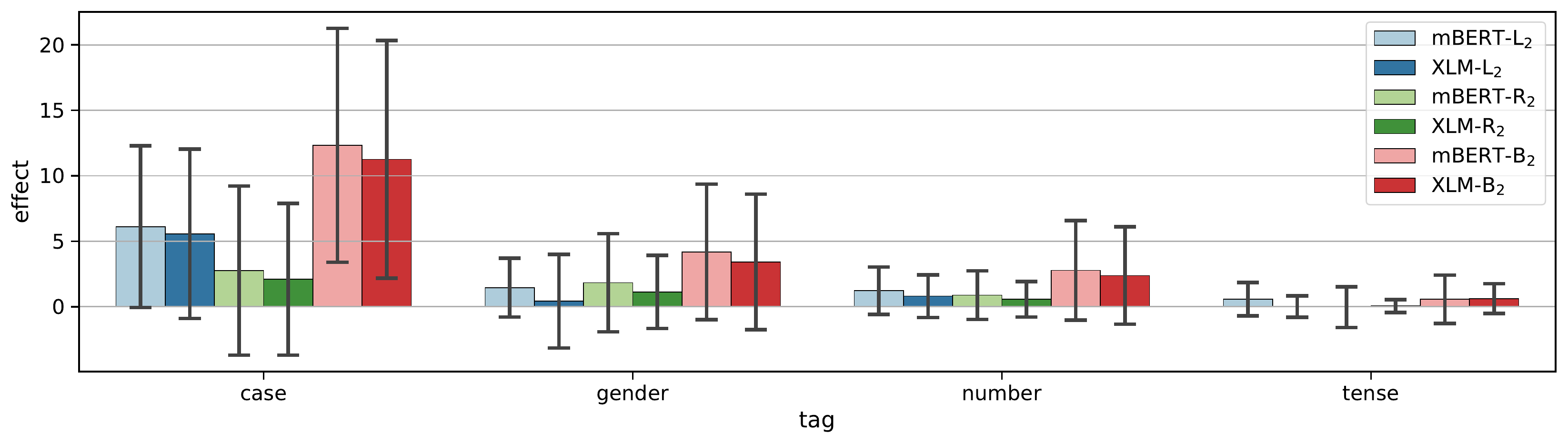}
    \vspace{0.5cm}
    \caption{The effect of context masking perturbations by tag. Error bars
    indicate the standard deviation.}
\label{fig:context_masking_by_tag} 
\end{figure}

Although the average effect is rather modest, there are some tasks that are
affected significantly by context masking perturbations.
\figref{context_masking_by_tag} shows the effect (as defined in
\equref{perturbation_effect}) by tag.

\begin{figure}
    \centering
    \includegraphics[clip,width=0.99\textwidth]{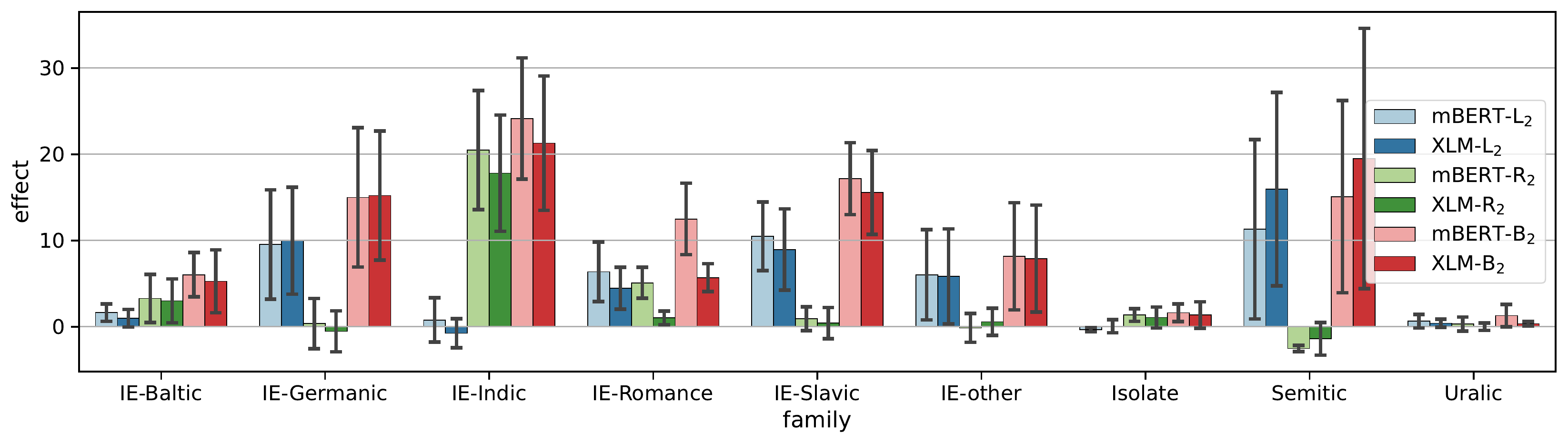}
    \vspace{0.5cm}
    \caption{The effect of context masking on case tasks groupby by language
    family. Error bars indicate the standard deviation.}
\label{fig:context_masking_case_by_family} 
\end{figure}

Since case is affected the most, we examine it a little closer.
\figref{context_masking_case_by_family} shows the effect of context masking on
case tasks grouped by language family. Uralic results are barely affected by
context masking, which confirms that the target word alone is indicative of the
case in Uralic languages. Germanic, Semitic, and Slavic case probes are
moderately affected by \pLtwo and somewhat surprisingly, we find a small
improvement in probing accuracy, by \pRtwo.  Indic probes are the opposite,
\pRtwo has over 20\% effect, while \pLtwo is close to 0. Indic word order is
quite complex, with a basic SOV word order affected both by split ergativity
and communicative dynamism (topic/focus) effects \citep{Jawaid:2011}. Again we
suspect that these complexities overwhelm the MLMs, which work best with
mountains of data, typically multi-gigaword corpora, 3-4 orders of magnitude
more than what can reasonably be expected from primary linguistic data, less
than thirty million words during language acquisition \citep{Hart:1995}.

\subsubsection{Target Masking and Word Order}\label{ssec:targ_permute}

\begin{figure}
    \centering
    \includegraphics[width=\textwidth]{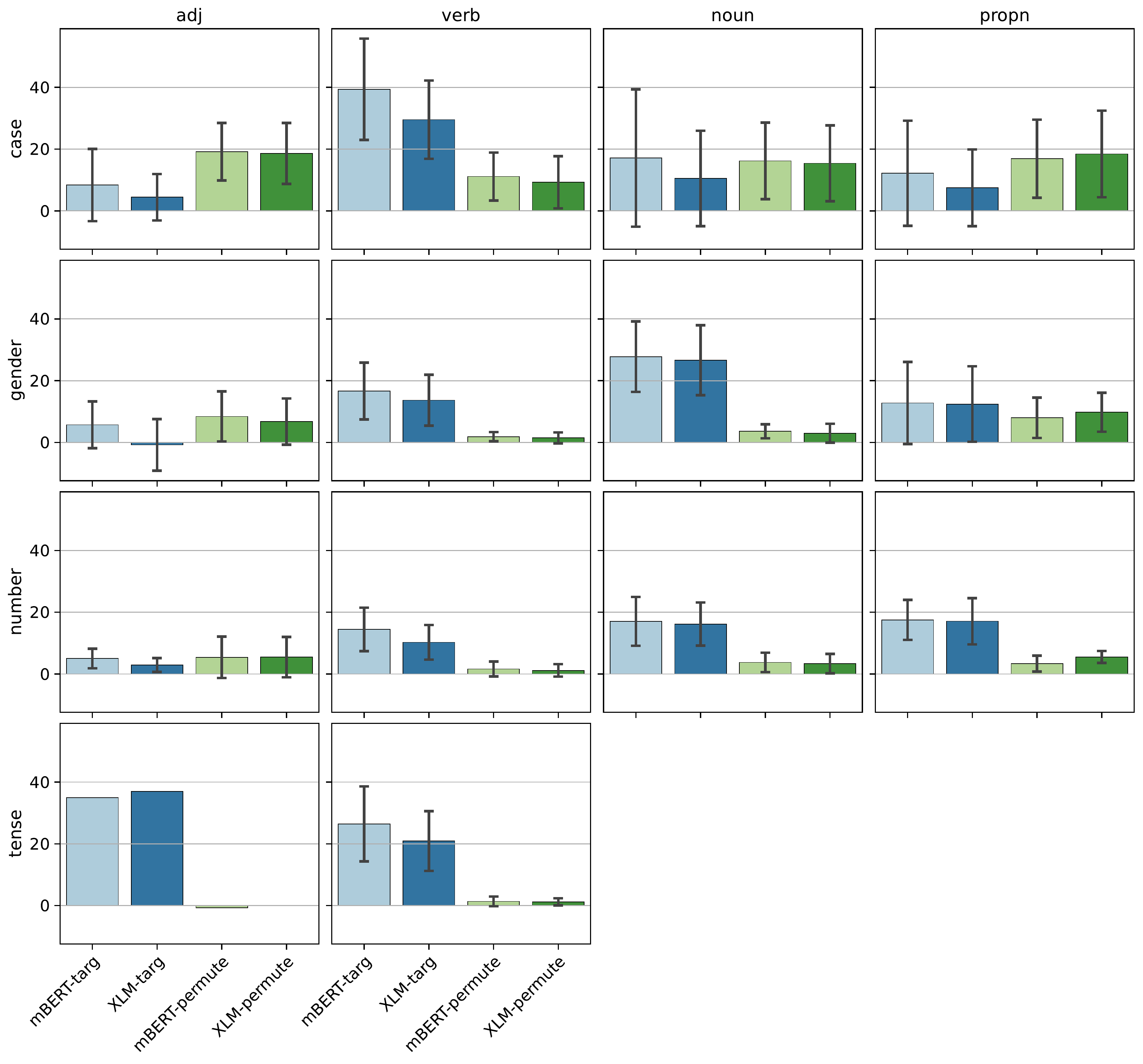}
    \vspace{0.2cm}
    \caption{The effect of \pTARG and \ppermute. Error bars indicate the standard deviation.}
    \label{fig:4x4_targ_permute}
\end{figure}

We discuss \pTARG and \ppermute in conjunction since they often have inverse
effect for certain languages and language families.  Target masking or \pTARG
is by far the most destructive perturbation with an average effect of 16.1\%
for \mbert and 12.7\% for \xlm. \ppermute is also a significant perturbation,
particularly for case tasks and adjectives.  As \figref{4x4_targ_permute}
shows, the effects differ widely among tasks but some trends are clearly
visible. \pTARG clearly plays an important role in many if not all tasks.
Verbal tasks rely almost exclusively on the target form and \ppermute has
little to no effect. Verbal morphology is most often marked on the verb form
itself, so this not surprising. Nouns and proper nouns behave similarly with
the exception of case tasks. Case tasks show a mixed picture for all 4 parts of
speech. \pTARG and \ppermute both have a moderate effect.  This might be
explained by the fact that case is expressed in two distinct ways depending on
the language. Agglutinative languages express case through suffixes, while
analytic languages, such as English, express case with prepositions. In other
words, the context is unnecessary for the first group and indispensable for the
second. 

Both \pTARG and \ppermute are markedly small for gender and number tasks in
adjectives. This is likely due to the fact that adjectives do not determine the
gender or the number of the nominal head but rather copy (agree with) it.

\begin{figure}
    \centering
    \includegraphics[width=\textwidth]{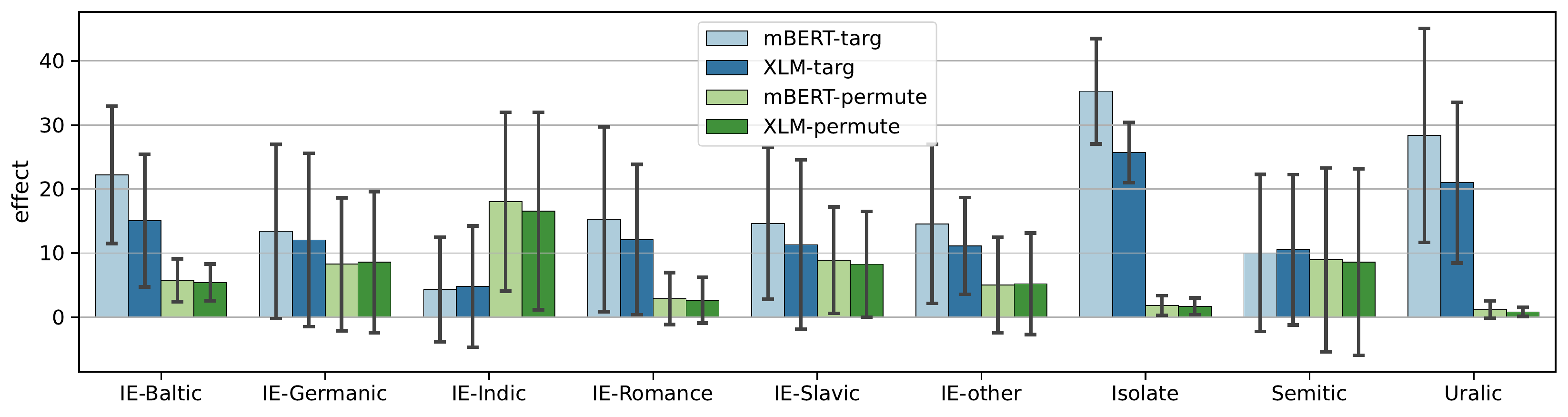}
    \vspace{0.5cm}
    \caption{The effect of \pTARG and \ppermute by language family. Error bars indicate the standard deviation.}
    \label{fig:targ_permute_by_family}
\end{figure}

\figref{targ_permute_by_family} shows the effect of \pTARG and \ppermute by
language family. Although the standard deviations are often larger than the
mean effects, the trends are clear for multiple language families. The Uralic
family is barely affected by \ppermute while \pTARG has over 20\% effect for
both models. \pTARG has a larger effect than \ppermute for the Baltic and the
Romance family and isolate languages. Indic tasks on the other hand tend to
have little change due to \pTARG, while \ppermute has the largest effect for
this family.

\subsubsection{Relationship between Perturbations}

\begin{figure}
    \centering
    \includegraphics[width=0.5\textwidth]{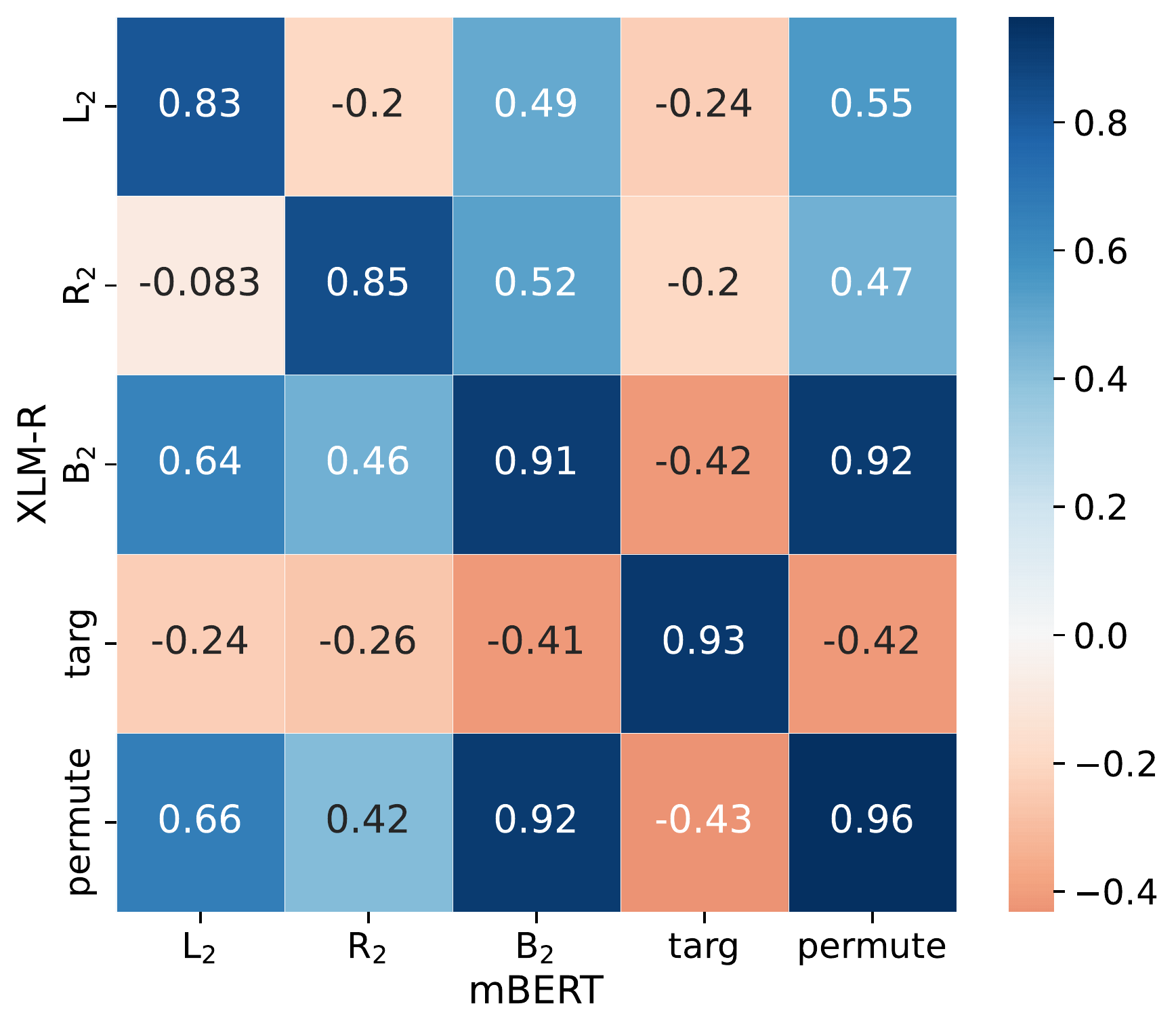}
    \vspace{0.5cm}
    \caption{The pairwise Pearson correlation of perturbation effects between
    the two models.}
    \label{fig:model_pert_corr}
\end{figure}

\begin{figure}
    \centering
    \includegraphics[width=\textwidth]{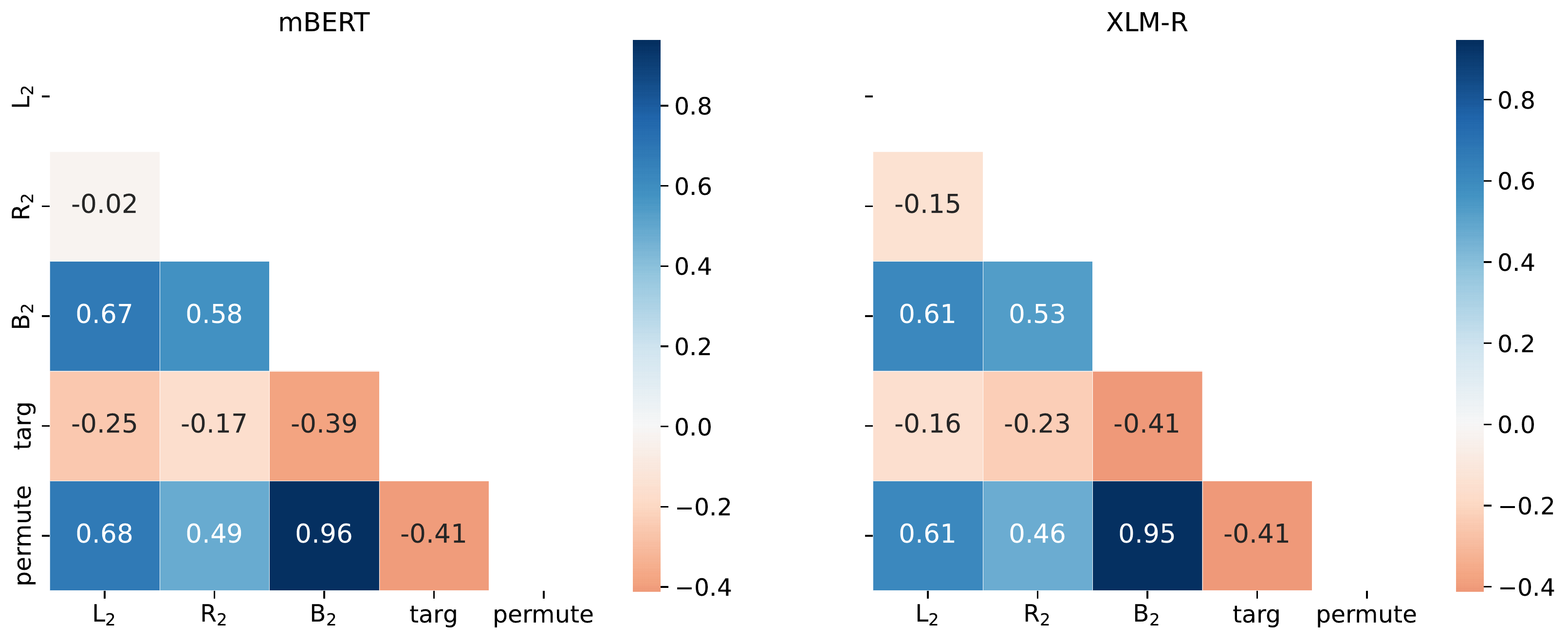}
    \vspace{0.5cm}
    \caption{The pairwise Pearson correlation of perturbation effects by model.}
    \label{fig:pert_corr_by_model}
\end{figure}

In the previous section we showed that \pTARG and \ppermute often have an
inverse correlation. Here we quantify their relationship as well as the
relationship between all perturbations across the two models.  First, we show
that the effects across models are highly correlated as evidenced by
\figref{model_pert_corr}, which shows the pairwise Pearson correlation of the
effects of each perturbation pair. The matrix is almost symmetrical. The main
diagonal is close to one, which means that the same perturbation affects the
two models in a very similar way. This suggests not just that the models are
quite similar (see also \figref{pert_corr_by_model} depicting the correlation
between perturbations in each model side by side) but also that the
perturbations tell us more about morphology than about the models themselves.

\subsection{Typology}

\begin{figure}
    \includegraphics[width=.99\textwidth,clip,trim={0 -1.7cm 0 0}]{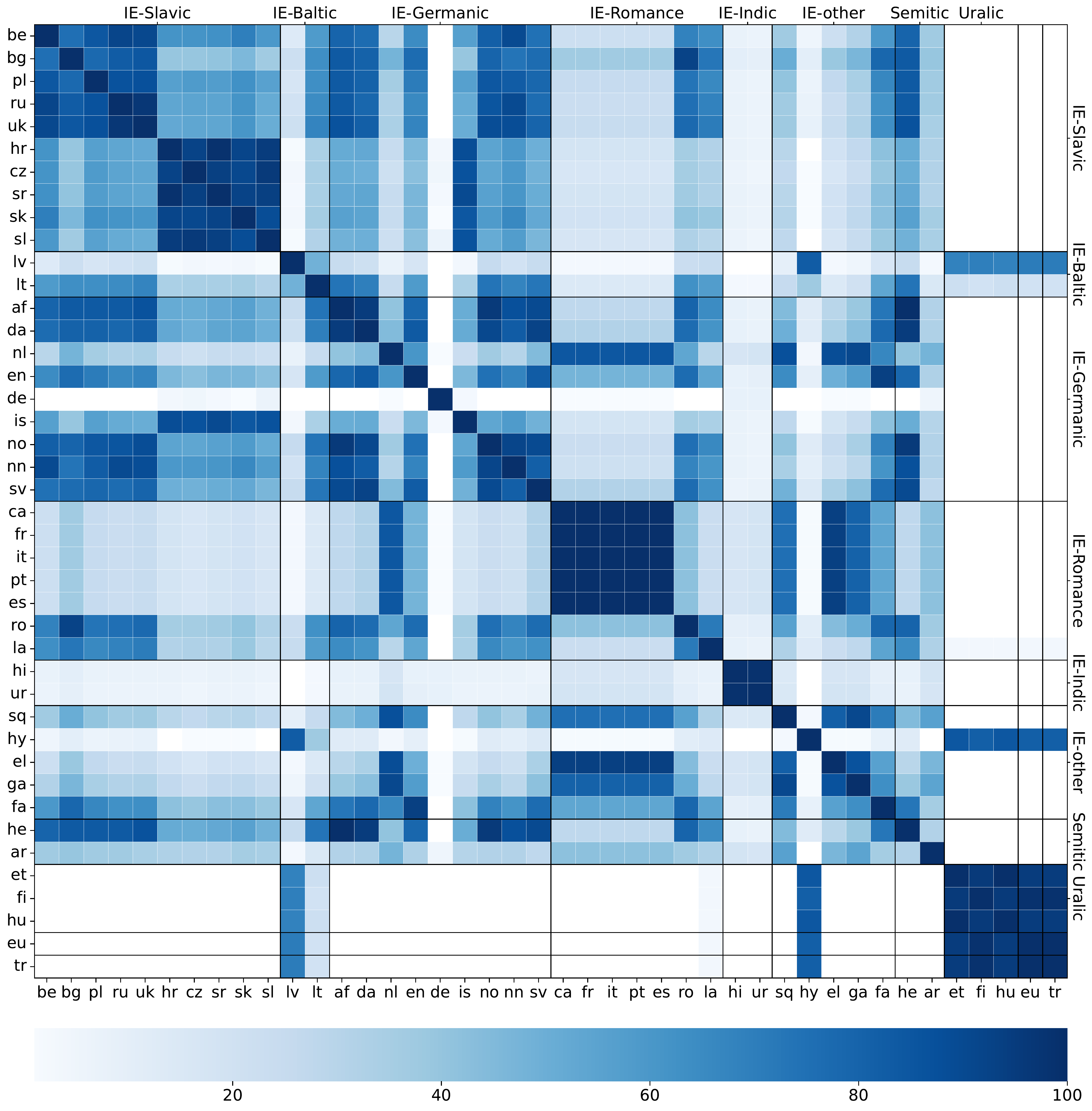}
    \caption{Co-occurrence counts for each languages pair
    over 100 clustering runs. Languages are sorted by family and a line is
    added between families.}
    \label{fig:lang_clustering}
\end{figure}

While our dataset is too small for drawing far-reaching conclusions, we are
beginning to see an emerging typological pattern in the effects of perturbation
as defined in \equref{perturbation_effect}. We cluster the languages
by the effects of the perturbations on each task. There are 5 perturbations and
\postagcombination tasks, available as input features for the clustering
algorithm, but many are missing in most languages. We use the column averages
as imputation values.  Since a single clustering run shows highly unstable
results, we aggregate over 100 runs of $K$-means clustering with $K$ drawn
uniformly between 3 and 8 clusters. We then count how many times each pair of
languages were clustered into the same cluster.  \figref{lang_clustering}
illustrates the co-occurrence counts for \xlm.  Since \mbert results are very
similar, we limit our analysis to \xlm for simplicity.

Language families tend to be clustered together with some notable exceptions.
German is seldom clustered together with other languages, including other
members of the Germanic family, except perhaps for Icelandic. To a lesser
extent, Latin is an outlier in the Romance family -- it clusters better with
Romanian than with Western or Southern Romance.  The two Indic languages are
almost always in a single cluster without any other languages, but the two
Semitic languages are almost never in the same cluster.  Arabic tends to be in
its own cluster, while Hebrew is often grouped with Indo European languages.
The Uralic family forms a strong cluster along with Basque and Turkish. These
languages have highly complex agglutination and they all lack gender, so this
is not surprising.

\section{Shapley Values}
\label{sec:shapley}

Having measured the (generally harmful) effect of perturbations, our next goal
is to assign responsibility (blame) to the contributing factors. We use Shapley
values for this purpose.  For a general introduction, see \citet{Shapley:1951}
and \citet{Lundberg:2017}; for motivation of Shapley values in NLP, see
\citet{Ethayarajh:2021}. We consider a probe as a coalition game of the words
of the sentence. We treat each token position as a player in the game. The
tokens are defined by their relative position to the target token. A sentence
is a sequence defined as $ L_k, L_{k-1}, \dots, L_1, T, R_1, R_2, \dots,
R_{m}$, where $k$ is the number of words that precede that target word and $m$
is the number of words that follow it. The tokens far to the left are
considered as belonging to a single position ($-4^-$), those far to the right
to another position ($4^+$), so we have a total of 9 players
$N=\{-4^-,-3,-2,-1,0,1,2,3,4^+\}$. 
On a given task, we can remove the contribution of a player $i$ by masking the
word(s) in positions corresponding to that player. The Shapley value
$\varphi(i)$ corresponding to this player is computed as

\begin{equation}
\varphi(i) = \frac{1}{n} \sum_{S \subseteq N \setminus \{i\}} \frac{v(S \cup
  \{i\}) - v(S)}{\binom{n-1}{|S|}},
\label{eq:shapley}
\end{equation}

\noindent where $n$ is the total number of players, 9 in our case, and $v(S)$
is the value of coalition $S$ (a set of players, here positions) on the given
tasks.  $v(S)$ is a function of the accuracies ($\mathrm{Acc}$) of the task's
probe with coalition $S$, the full set of players $N$, and the model. When all
players are absent (masked), $\mathrm{Acc}_\text{all masked}$ is very close to
the accuracy of the trivial classifier that always picks the most common label.
As is clear from Eq.~\ref{eq:shapley}, the contribution of the $i$th player is
established as a weighted sum of the difference in the contributions of each
coalition that contains $i$ versus having $i$ excluded. The weights are chosen to
guarantee that these contributions are always additive: bringing players $i$
and $j$ into a coalition improves it exactly by $\varphi(i)+\varphi(j)$. The
value of the entire set of players is always 1 (we use a multiplier 100 to
report results in percentages), and we scale the contributions so that the
value of the empty coalition is 0:

\begin{equation}
    v(S) = 100 - 100 \cdot \frac{\mathrm{Acc}_S - \mathrm{Acc}_\text{all
    masked}}{\mathrm{Acc}_{\text{\mbert}} - \mathrm{Acc}_\text{all masked}}.
\label{eq:scaled_effect}
\end{equation}

\noindent
Not only are the Shapley values defined by Equation~\ref{eq:shapley} an
additive measure of the contributions that a particular player (in our case,
the average word occurring in that position) makes to solving the task, but
they define the only such measure \citep{Shapley:1951}.

\subsection{Implementation}

Both \mbert and \xlm have built-in mask tokens that are used for the masked
language modeling objective. We remove the contribution of certain tokens by
replacing them with mask symbols. Multiple tokens can be removed at a time and
we use a single mask token in place of each token. We designate an unused
character as mask for the chLSTM experiments. When a token is masked, we
replace each of its characters with this mask token. Computing the Shapley
values for 9 players requires $2^9=512$ experiments for each of the \taskno
tasks. This includes the unmasked sentence (all players contribute) and the
completely masked sentence (no players), where each token is replaced with a
mask symbol. 

\subsection{General Results}

\begin{table}
    \centering
    \caption{Summary of the Shapley values.}
    {\tablefont
\begin{tabular}{lrrrrrrr}
\hline
 model &  left &  right &  target &  context &  left/right &  left/context &  right/context \\
\hline
 mBERT &  24.3 &   16.7 &    58.9 &     41.1 &       1.455 &         0.591 &          0.406 \\\hdashline
 XLM-R &  27.0 &   18.6 &    54.4 &     45.6 &       1.452 &         0.592 &          0.408 \\\hdashline
chLSTM &  15.7 &    5.3 &    79.0 &     21.0 &       2.962 &         0.748 &          0.252 \\
\hline
\end{tabular}
    }
    \label{tab:shapley_summary} 
\end{table}

\begin{figure}
    \centering
    \includegraphics[width=0.6\textwidth]{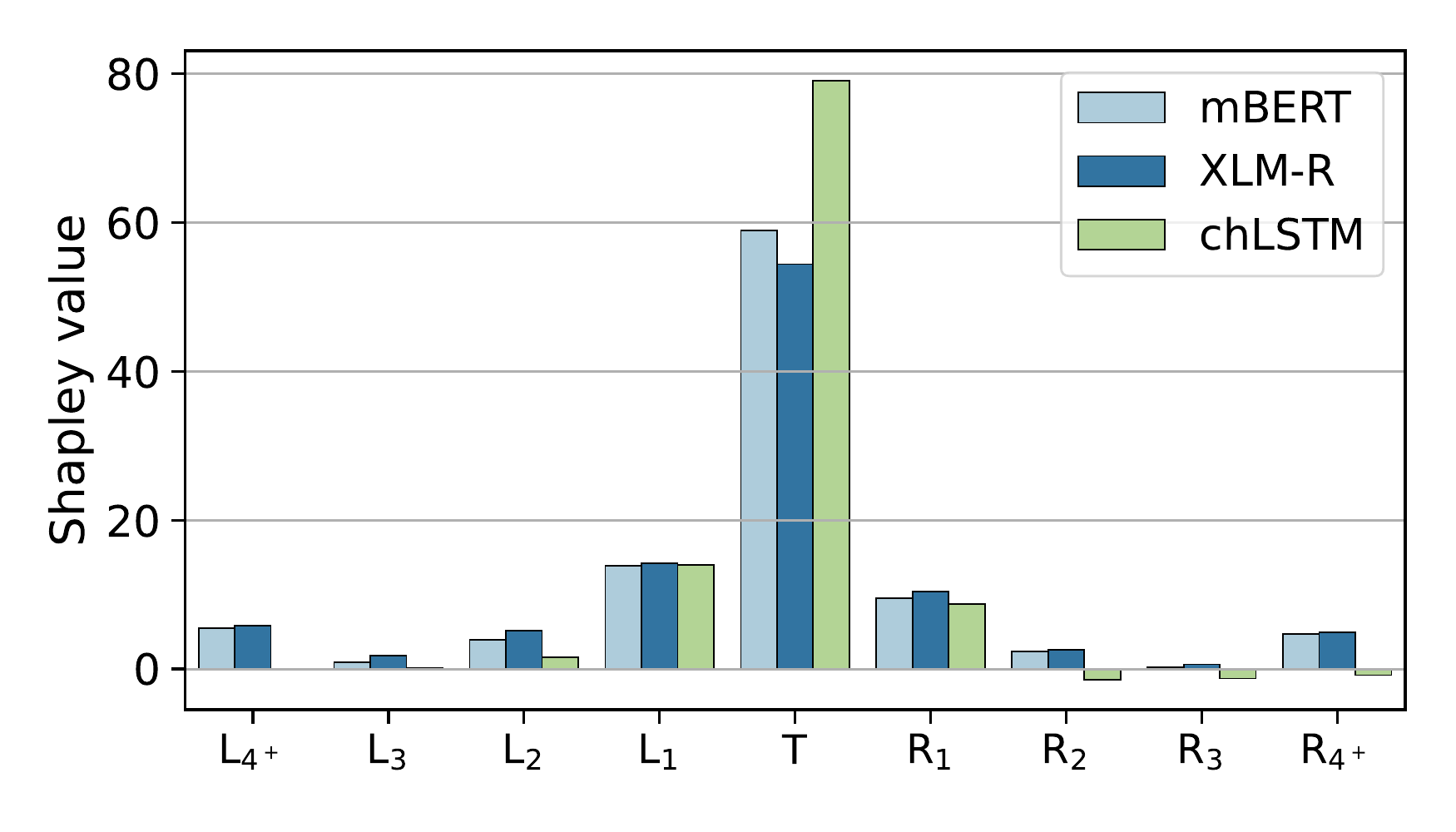}
    \vspace{1cm}
    \caption{Shapley values by relative position to the probed target word. The
    values are averaged over the \taskno tasks.}
    \label{fig:shapley_average}
\end{figure}

\figref{shapley_average} shows the Shapley values averaged over the \taskno
tasks for each model. \tabref{shapley_summary} summarizes the numerical
results. The values extracted from the two MLMs are remarkably similar.  We
quantify this similarity using $L_1$ (Manhattan) distance, which is 0.09
between the means.\footnote{The Kullback-Leibler (KL) divergence is 0.014 bits,
also very small, but we use $L_1$ in these comparisons, since individual
Shapley values can be negative. $L_2$ (Euclidean) distance values would be
just as good (the Pearson correlation between $L_1$ and $L_2$ is 0.983), but
since Shapley values sum to 1 Manhattan is easier to interpret. In what
follows, ``distance'' always refers to $L_1$ distance.} The Shapley
distributions obtained by \xlm and \mbert move closely together: the mean
distance between Shapley values obtained from \xlm and \mbert is just 0.206,
and of the \taskno pairwise comparisons only 5 are more than two standard
deviations above the mean. This means that in general Shapley values are more
specific to the morphology of the language than to the model we probe. To
simplify our analysis, we only discuss the \xlm results in detail since they
show the same tendencies and are slightly better than the results achieved with
\mbert.

The first observation is that the majority of the information, 54.9\%, comes
from the target words themselves, with the context contributing on average
only 45.1\%. Next, we observe that words further away from the target
contribute less, providing a window weighting scheme (kernel density function)
broadly analogous to the windowing schemes used in speech processing
\citep{Harris:1978b}. Third, the low Shapley values at the two ends, summing
to 11.2\% in \xlm (11.0\% in \mbert) go some way toward vindicating the
standard practice in KWIC indexing \citep{Luhn:1959}, which is to retain only
three words on each side of the target. While the observation that this much
context is sufficient for most purposes, including disambiguation and machine
translation, goes back to the very beginnings of information retrieval (IR) and
machine translation (MT) \citep{Choueka:1985}, our findings provide the first
quantifiable statement to this effect in MLMs (for HMMs, see
\citealp{Sharan:2018}) and open the way for further systematic study directly on
IR and MT downstream tasks.

With this we are coming to our central observation, evident both from
\figref{shapley_average} and from numerical considerations
(\tabref{shapley_summary}): the decline is
noticeably faster to the right than to the left, in spite of the fact that
there is nothing in the model architecture to cause such an asymmetry (see
\ref{ssec:directionality}). What is more, not even our experiments with random 
weighted MLMs (presented in \ref{ssec:randinit}) show such asymmetry.

Whatever happens before a target word is about
40\% more relevant than whatever happens after it. In morphophonology
`assimilation' is standardly classified, depending on the direction of
influence in a sequence, as {\it progressive} assimilation, in which a
following element adapts itself to a preceding one, and {\it regressive} (or
anticipatory) assimilation, in which a preceding element takes on a feature or
features of a following one. What the Shapley values suggest for
morphology is that progressive assimilation (feature spreading) is more
relevant than regressive. 

This is not to say that regressive assimilation will be impossible, or even
rare. One can perfectly well imagine a language where adjectives precede the
noun they modify and agree to them in gender:\footnote{Indeed, there are
several such languages in our sample such as German and most Slavic
languages.}  this form of agreement is clearly anticipatory. Also, the
direction of the spreading may depend more on structural position than linear
order, cf. for example the `head marking' versus `dependent marking'
distinction drawn by \cite{Nichols:1986}. But when all is said and done, the
Shapley values, having been obtained from models that are perfectly
directionless, speak for themselves: left context dominates right 58.39\% to
41.61\% in \xlm (58.36\% to 41.64\% in \mbert) when context weights are
considered 100\%. This makes clear that it is progressive, rather than
anticipatory, feature sharing that is the unmarked case. While our dataset is
currently heavily skewed toward IE languages, so the result may not hold on a
typologically more balanced sample, it is worth noting that the IE family is
very broad typologically, and three of the four heaviest outliers (Hindi,
Urdu, Irish) are from IE, only Arabic is not.

\subsection{Outliers}\label{ss:outliers}

\begin{figure}
    \centering
    \includegraphics[clip,trim={0 0 0
    0cm},width=\textwidth]{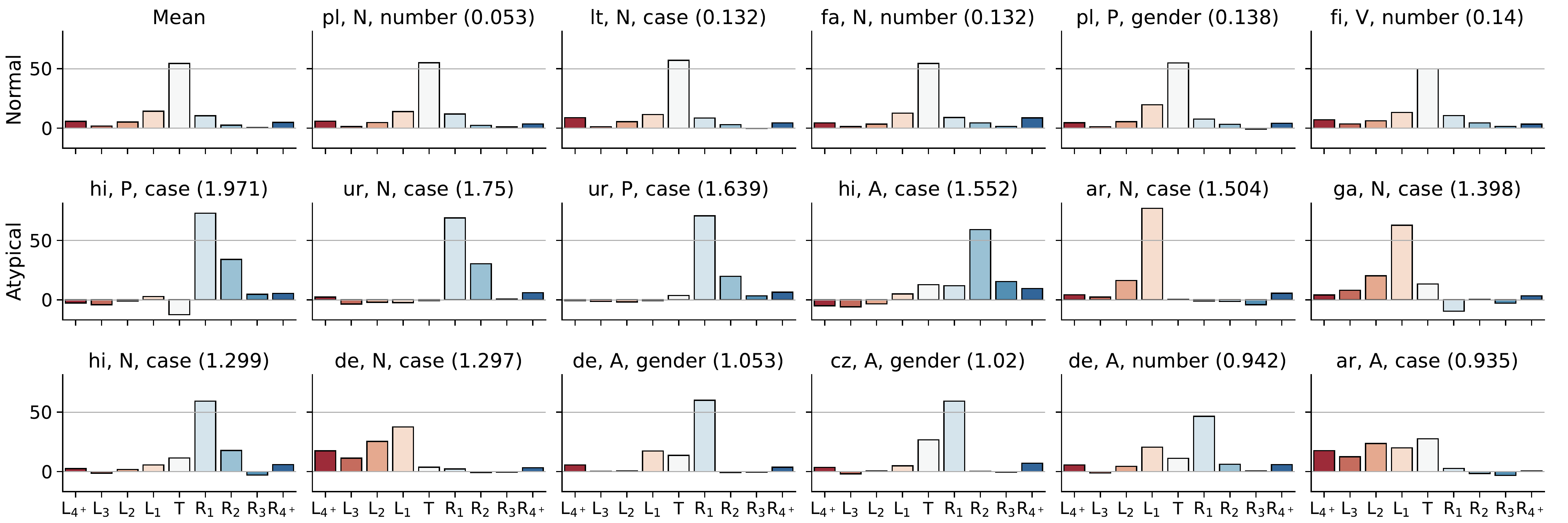}
    \vspace{1.5cm}

    \caption{Least and most anomalous Shapley distributions. The first row are
      the mean Shapley values of the \taskno tasks and the 5 tasks
      \emph{closest} to the mean distribution, i.e.~the least anomalous as
      measured by the dfm distance from the average Shapley values. The rest
      of the rows are the most anomalous Shapley values in descending
      order. For each particular task, its distance from the mean (dfm) is
      listed in parentheses above the graphs. }

    \label{fig:shapley_outliers} 
\end{figure}

We next consider the outliers. The main outliers are listed in
\figref{shapley_outliers}. We compute the distance of each task's Shapley
values from the mean (dfm). Over 91.5\% of the tasks are very close (Manhattan
distance below one standard deviation, 0.264) to the mean of the distribution, and there are
only 5 tasks (2\% of the total) where the distance exceeds two standard deviations above
the mean.  The first row of \figref{shapley_outliers} shows the mean
distribution and the 5 tasks that are {\it closest} to it, such as
\task{Polish}{N}{number} (1st row 2nd panel, distance from mean 0.053) or
\task{Lithuanian}{N}{case} (1st row 3rd panel, dfm 0.132).  These exemplify
the typologically least marked, simplest cases, and thus require no special
explanation.

\begin{figure}
    \centering
    \includegraphics[clip,trim={0 0 0 0cm},width=.98\textwidth]{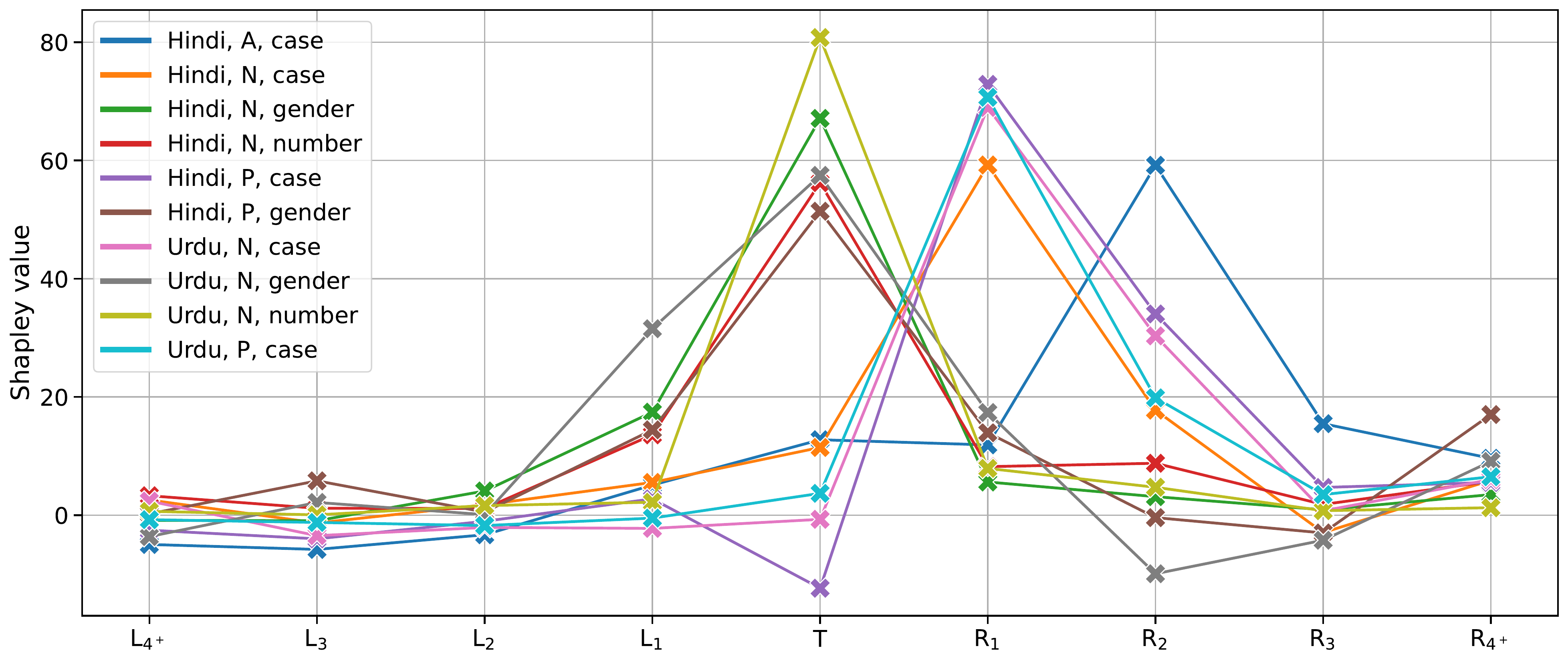}
    \vspace{0.3cm}
    \caption{Shapley values in Indic tasks.}
    \label{fig:shapley_indic}
\end{figure}

What does require explanation are the outliers, Shapley patterns far away from
the norm. By distance from the mean, the biggest outliers are Indic:
\task{Hindi}{PROPN}{case} and \task{Hindi}{ADJ}{case} (2rd row 1st and 4th panels,
dfm 1.971 and 1.552 respectively) and \task{Urdu}{NOUN}{case} and
\task{Urdu}{PROPN}{case} (2nd row 2nd and 3rd panels, dfm 1.75 and 1.639
respectively), see \figref{shapley_indic}. For proper nouns, the greatest
Shapley contribution, about 72--73\%, is on the word following the proper
noun. In Hindi not knowing the target is actually better than knowing it, the
target’s own contribution is negative 12\% (and in Urdu, a minuscule 3\%). For
the case marked on Hindi adjectives, the most important is the second to its
right, 59\%; followed by the 3rd to the right, 15\%; the target itself, 13\%;
and the first to the right, 12\% (we do not have sufficient data for Urdu
adjectives).  The Indic noun case patterns, unsurprisingly, follow closely the
proper noun patterns. For both Hindi and Urdu there are good typological
reasons, SOV word order, for this to be so.\footnote{We thank Paul Kiparsky
(pc) for pointing this out.}

The next biggest outliers are \task{Arabic}{NOUN}{case} and
\task{Irish}{NOUN}{case} (\figref{shapley_outliers} 2nd row 5th and 6th panel,
dfm 1.505 resp. 1.398), where the preceding word is more informative than the
target itself. These are similarly explainable, this time by VSO order. It also
stands to reason that the preceeding word, typically an article, will be more
informative about \task{German}{NOUN}{case} than the word itself (3nd row 2rd
panel, dfm 1.297). The same can be said about \task{German}{ADJ}{gender} (dfm
1.053) and \task{German}{ADJ}{number} (dfm 0.942), or the fact that
\task{Czech}{ADJ}{gender} (3rd row 4th panel, dfm 1.02) is determined by the
following word, generally the head noun.

If we arrange Shapley distributions by decreasing distance from the mean, we
see that dfm is roughly normally distributed (mean 0.492, std 0.264). Only 21
tasks are more than one standard deviation above the mean, the last two rows
of \figref{shapley_outliers}, present the top 12 of these.  Altogether, there
was a single case where $R_2$ dominated, \task{Hindi}{ADJ}{case}, 16 cases when
$L_1$ dominates, and 11 cases where $R_1$ dominates, everywhere else it is the
target that is the most informative. The typologically unusual patterns, all
clearly related to the grammar of the language in question, are transparently
depicted in the Shapley patterns.  For example, as noted in \ref{ss:difftask},
the article preceding the noun in German often is the only indication of the
noun's case. The Shapley values we obtained simply quantify this information
dependence. Similarly, Arabic cases are determined in part by the preceding
verb and/or preposition. Quite often, Shapley values confirm what we know
anyway, e.g., that verbal tasks rely more on the target word than nominal
tasks.

Another noticeable statistical trace of rule-governed behavior is seen in
Hindi and Urdu, where the oblique case appears only
when governed by a postposition.  Therefore, the presence of a postposition in
$R_1$ is diagnostic for the case of a target noun, and its presence in $R_2$
is diagnostic for a target adjective.  This conclusion is confirmed by the
Shapley values, which are dominated by $R_1$ for case in Hindi nouns and
proper nouns (67.7\% and 64.9\% respectively) and by $R_2$ for Hindi
adjectives (82\%). Urdu noun and proper noun cases show the same $R_1$
dominance (64.4\% and 68.5\%). In contrast, the overall average Shapley value
of $R_1$ and $R_2$ are only 10.4\% and 2.9\% (see \figref{shapley_indic} for
the full patterns). 

To the extent that similar rule-based explanations can be ascertained for all
cases listed in \figref{shapley_outliers}, we can attribute \xlm's success to
an impressive sensitivity to grammatical regularities.  Though the mechanisms
are clearly different, such a finding places \xlm in the same broad tradition
as other works seeking to discover rules and constraints
\citep[e.g.,][]{Brill:1993}.

\subsection{The Difficulties of Generalization} \label{ssec:shapley_generalization}

Our Shapley data can be summarized as a 9-dimensional vector for each
\task{i}{j}{k}. In other words, the Shapley distributions come naturally
arranged in a 3D tensor. Unfortunately, many of the values are missing, either
because the language does not combine a particular POS with a particular tag, or
because we do not have enough data for training on the task. With a much larger
dataset (recall from \ref{ss:lgtag} that we use essentially all currently
available uniformly coded data) experimenting with 3D tensor decomposition
techniques \citep{Kolda:2009} may make sense, but for now the outcome depends
too much on the imputation method. That said, we have obtained one robust
conclusion, independent of how we fill the missing values: it is harder to
generalize by language than by POS or tag. It would be tempting to look at
languages as units of generalization, but we found that trends rarely apply to
individual languages! 

For the totality of tasks \task{i}{j}{k} we can keep one of $i,j$, or $k$
fixed, and compute the average Shapley distributions $S_{j,k}(i), S_{i,k}(j)$,
and $S_{i,j}(k)$. Given the average distributions, say $S_{j,k}$(Polish) we can
ask how far Shapley distributions for all available Polish tasks are from it,
and compute the average of these distances from the mean in the selected
direction (in this case, language). We find that the average distance from the
language averages is 0.37, while the distance from tag averages is 0.25 and the
distance from POS averages is 0.26. In other words, aggregating tasks by
language results in considerably larger variability than aggregating by POS or
tag. The POS and tag results are similar since POS and tag are highly
predictive of each other across languages: typically nouns will have case,
verbs will have tense, and conversely, tenses are found on verbs, cases on
nouns. This makes data aggregated on POS and tag jointly, as in
\figref{4x4_targ_permute}, much easier to make sense than data aggregated by
language.

\section{Ablations} \label{sec:ablations}

In this section we empirically consider the criticisms raised in
\cite{Belinkov:2022} and \cite{Ravichander:2021} for probing setups like ours.
Our first group of tests (\ref{ssec:mlp_variations}) confirms that the probing
accuracy does not depend on the choice of probe, in particular, linear probes
are no better or worse than non-linear ones. We also show that probing
individual layers of \mbert or \xlm is worse or no better than probing the
weighted sum of all layers (\ref{ssec:layer_pooling}). We also show that
fine-tuning decreases the probing accuracy while it substantially increases the
computational requirements (\ref{ssec:finetuning}). Finally, we show that
probing a randomly initialized model, a control used by \citet{Voita:2020}, is
significantly worse than probing the trained model (\ref{ssec:randinit}). We
present all results averaged over all \taskno tasks in this section. With the
exception of \ref{ssec:randinit}, we do not perturb the input sentences.

\subsection{Linear Probing and MLP Variations}\label{ssec:mlp_variations}

The probes we presented so far all use an MLP with a single hidden layer with
50 neurons. The input is the weighted sum of the 12 layers and the embedding
layer with learned weights. The size of the output layer depends on the number of
classes in the probing tasks. We use ReLU activations in the MLP. 

The original BERT paper \citep{Devlin:2019} used a simple linear classification
layer with weight $W \in \mathbb{R}^{K \times H}$, where $K$ is the number of
labels and $H$ is the hidden size of BERT, 768 in the case of \mbert and \xlm.
\citet{Hewitt:2019b} argue that linear probes have high selectivity,
i.e., they
tend to memorize less than non-linear probes. We test this on our dataset with
two kinds of linear probes. The first one is the same as the general probing
setup but we remove the ReLU activation. The second one completely removes the
hidden layer similarly to the original BERT paper. We also test two MLP
variations, one with 100 hidden size instead of 50 and another one with two
hidden layers.

\begin{figure}
    \centering
    \vspace{0.3cm}
    \includegraphics[width=0.6\textwidth]{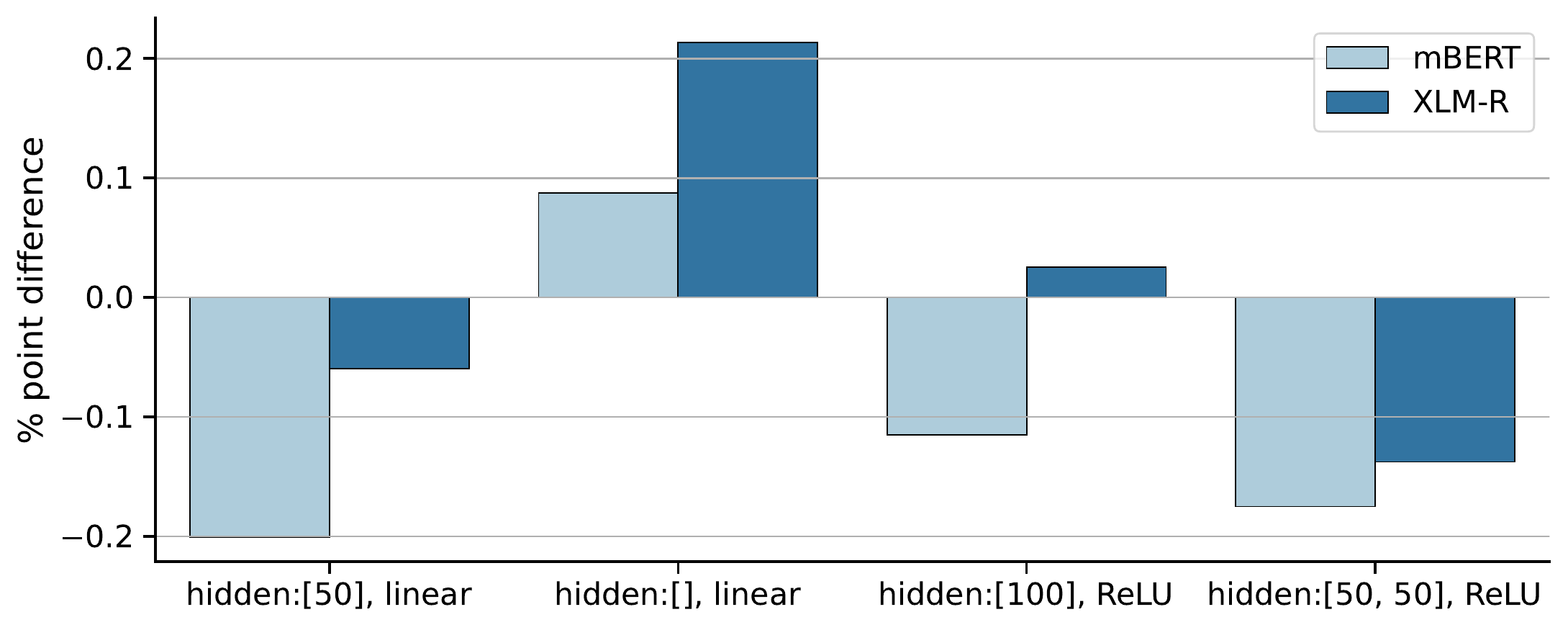}
    \vspace{0.8cm}
    \caption{The average probing accuracy using different MLP variations. We
    indicate the size(s) of hidden layer(s) in square brackets.}
    \label{fig:mlp_variations}
\end{figure}

\figref{mlp_variations} shows the accuracy of the two linear probes and the
larger MLPs as the difference from the default version we used
elsewhere. These numbers are averaged over that \taskno tasks. The differences
are all smaller than 0.25\% points. These results indicate that the
probing accuracy does not depend on the probe type and particularly that linear probes 
perform similarly to non-linear ones.

\subsection{Layer Pooling}\label{ssec:layer_pooling}

Our default setup uses a weighted sum of the 12 layers and the embedding
layers, one scalar weight for each of them with a total of 13 learned weights.
It has been shown \citep{Tenney:2019a} that the different layers of \mbert work
better for different tasks. Lower layers work better for low level tasks such
as POS tagging, while higher layers work better for higher level tasks such as
coreference resolution. Morphosyntactic tagging is a low level task. The
embedding layer itself is often indicative of the morphological role of a
token. We test this by probing each layer separately as well as probing the
concatenation of all layers.

\begin{figure}
    \centering
    \includegraphics[width=0.99\textwidth]{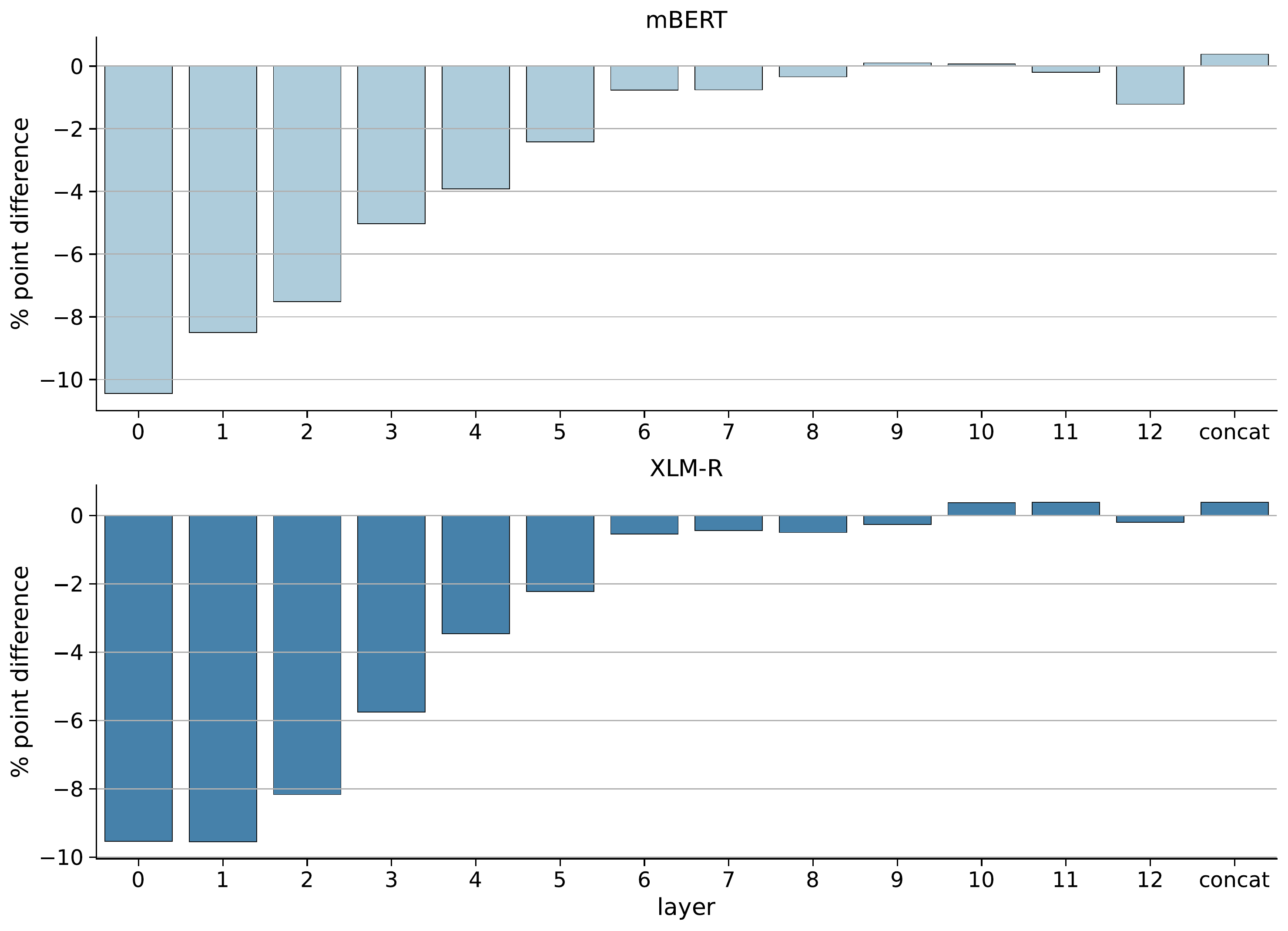}
    \vspace{1.8cm}

    \caption{The difference between probing a single layer and probing the
        weighted sum of layers. \textit{concat} is the concatenation of all
        layers. \textit{0} is the embedding layer. Large negative values on the
        y-axis mean that probing the particular layer on the x-axis is much
    worse than probing the weighted sum of all layers.}

    \label{fig:layerwise_probing}
\end{figure}

\figref{layerwise_probing} shows the difference between probing the weighted
sum of all layers and probing individual layers averaged over all tasks.  We
observe approximately 10\% point difference in the embedding layer (layer 0)
and the lower layers. This difference gradually decreases and it is close to 0
in the upper layers. Our results support the finding of \citet{Hewitt:2021}
that morphosyntactic cues are encoded much higher into the layers\footnote{This
holds for not just \xlm, but \mbert as well.} then previously suggested by
\citet{Tenney:2019a}, discussed in \citet{Hewitt:2019b}. Layer
concatenation (concat) is slightly better than the weighted sum of the
layers, but it should be noted that the parameter count of the MLP is an
order of magnitude larger thanks to the 13 times larger input dimension.

We also observe that the gap between the embedding layer (layer 0) and the
first Transformer layer (layer 1) is much smaller in the case of \xlm than
\mbert. \xlm's embedding layer is significantly better than \mbert's embedding
layer to begin with (82.2\% vs.~80\%) and this gap shrinks to 0.3\% point at
the first layer (82.2\% and 81.9\%). This is more evidence for one of our main
observations that \xlm's embedding and vocabulary are better than those of
\mbert's.

\subsection{Fine-tuning}\label{ssec:finetuning}

Fine-tuning (as opposed to feature extraction), trains all BERT parameters
along with the MLP in an end-to-end fashion. This raises the number of
trainable parameters from 40k to over 170M.  The recommended optimizer for
fine-tuning BERT models is AdamW \citep{Loshchilov:2018}, a variant of the Adam
optimizer. We try both Adam and AdamW for fine-tuning the model on each task
and show that AdamW is indeed a better choice than Adam (\tabref{finetuning})
but nevertheless the feature extraction results are 0.5 percentage point better
than the fine-tuning results and this difference is statistically significant.
Our experiments also show that the running time is increased 80-fold when we
fine-tune \mbert.  Due to this increase in computation time, we do not repeat
the experiments for \xlm. It should be noted that BERT fine-tuning has its own
tricks \citep{Houlsby:2019,Li:2021,Ben-Zaken:2022} that may lead to better
results but we do not explore them in this paper.

\begin{table}
    \centering
    \caption{Comparison of fine-tuned and frozen (feature extraction) models.}
    {\tablefont
        \begin{tabular}{@{\extracolsep{\fill}}lrr}
            \hline
                & Adam & AdamW \\
                \hline
            Finetuning & 76.2 & 89.2 \\
            \hdashline
            Feature extraction & 90.4 & 89.9 \\
            \hline
        \end{tabular}
        \vspace{.5cm}
    }
    \label{tab:finetuning} 
\end{table}

\subsection{Randomly Initialized MLMs}\label{ssec:randinit} 

Randomly initialized language models have been widely used as a baseline when evaluating
language models \citep{Conneau:2017}, especially via auxiliary classifiers 
\citep{Conneau:2018b, Zhang:2018, Htut:2019, Voita:2020}. \citet{Zhang:2018} 
showed that the mechanism of assigning morphosyntactic tags to these random 
embeddings is significantly different. As they demonstrated, randomly 
initialized MLMs rely on word identities, while their trained counterparts 
maintain more abstract representations of the tokens in the input layer. 
Therefore, probing classifiers applied on random MLMs may pick up low-level patterns
only, such as word identity, and this could mislead the probing controls when used as a baseline. 

In order to test this hypothesis, we trained the probing classifiers using
randomly initialized \mbert and \xlm models. In this setup both fully random
and pre-trained embedding layer with random Transformer layers were compared to
trained MLMs.

\tabref{random_bert} shows the overall probing accuracy achieved on random
models. We add the majority (most frequent) baseline as a comparison. Although
the random MLMs are clearly better than the majority baseline, they are far worse
than the trained MLMs.

\begin{table}
    \centering
    \caption{Probing accuracy on the randomly initialized \mbert and \xlm models.}
    {\tablefont
        \begin{tabular}{@{\extracolsep{\fill}}lrrrr}
            \hline
                & Pre-trained & Random layers & Fully random & Majority baseline \\
                \hline
            \mbert & 90.4 & 73.5 & 67.0 & 41.0 \\
            \hdashline
            \xlm & 91.8 & 74.7 & 70.1 & 41.0 \\
            \hline
        \end{tabular}
    }
    \label{tab:random_bert} 
\end{table}

\begin{figure}
    \centering
    \includegraphics[width=0.99\textwidth]{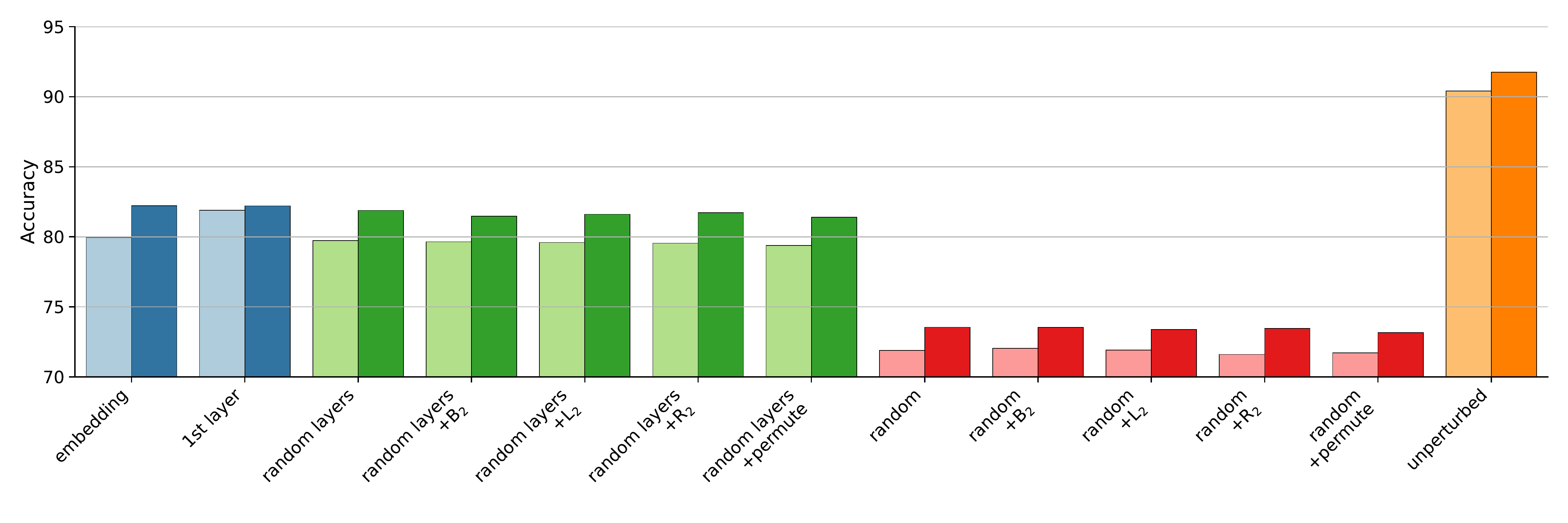}
    \vspace{1.7cm}
    \caption{
        Random \mbert (light color) and random \xlm (darker color) performance comparison with different 
        perturbation setups and the unperturbed trained model variants (orange bars). 
	Left-to-right:
        Blue: Accuracy of the embedding and first layers' probes;
        Green: Random models with pre-trained embedding layer: no perturbation, \pBtwo, \pLtwo, \pRtwo, \ppermute; 
        Red: Random models where the embedding layer is random as well: no perturbation, \pBtwo, \pLtwo, \pRtwo, \ppermute; 
        Orange: Unperturbed trained models.}
    \label{fig:rand_clms}
\end{figure}

\figref{rand_clms} shows that neither \pBtwo, \pLtwo, \pRtwo nor \ppermute
perturbations affect the model's performance in a way that they do when applied
on the trained models (compare \figref{perturbation_by_pos}).
Nevertheless, this sub-experiment offers two further supporting arguments for
the claims \citet{Zhang:2018} made about random MLMs learning word-identities:

\begin{enumerate}

    \item{The accuracies of random models' morphological probes match the
        accuracies of their embedding layers' probe; i.e.~even the
    Transformer-based random MLMs rely \emph{mostly} on the word-identities
    represented by their embeddings}

    \item{Probing perturbed and unperturbed embeddings of random MLMs does not
        make a big difference (the accuracies of unperturbed and perturbed
    models' are less than 1\% apart). Clearly, word identities count the most,
    their order almost not at all.}

\end{enumerate}

Based on this finding we do not use randomized MLMs as
baselines.

\subsection{Training Data Size} \label{ssec:training_data_size}

Our sampling method (cf.~\ref{ssec:data_generation}) generates 2,000
training samples. Raising this number would remove many tasks from languages with
smaller UD treebanks. Probing methods on the other hand are supposed to test
the already existing linguistic knowledge in the model, therefore
probing tasks' training sets
should not be too large. In this section, we show that smaller training sizes
result in inferior probing accuracy. Our choice of 2,000 samples was a
practical upper limit that allowed for a large number of tasks from
mid-to-large sized UD treebanks.

\begin{figure}
    \centering
    \includegraphics[width=0.7\textwidth]{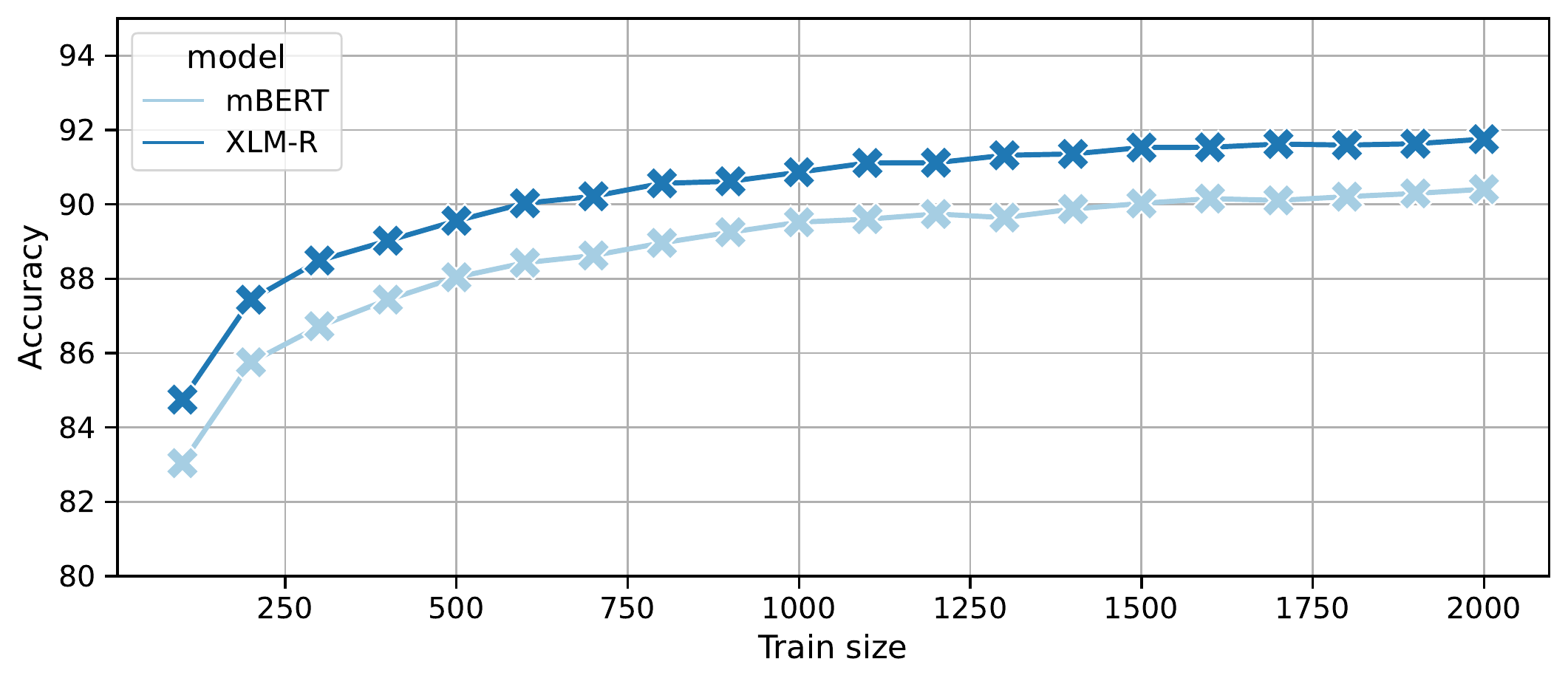}
    \vspace{.5cm}
    \caption{Probing accuracy with reduced training data.}
    \label{fig:train_size}
\end{figure}

\figref{train_size} shows the average accuracy of the
probing tasks when we use
fewer training samples. Although the probing tasks work considerably better
than the majority baseline even with 100 sentences, the overall accuracy gets
better as we increase the training data. Interestingly, \xlm is always slightly
better than \mbert.

\section{Discussion}\label{sec:discussions}

Throughout this paper we analyzed a large number of experiments and a few
trends were clear and corroborated by various experiments. In this section we
summarize these observations and point back to the experiments that confirm
them.

\paragraph{Pre-training learns morphology} Morphology is a low-level task and
for the most part it is considered an easy task as evidenced by the success of
small encoder-decoder networks in the SIGMORPHON shared tasks on token-level
morphology. Our strongest baseline, chLSTM, which is not pre-trained on
external data but only on the 2000 training sentences from the probing data,
performs reasonably well on most morphosyntactic tasks. We still find that the MLMs
and Stanza offer significant improvement,  which can only be attributed to the
their far larger training set (\tabref{task_model_averages}). chLSTM is
only better at 8 tasks out of the \taskno and the difference is never large.
The difference is not uniform across tags and POS
(Figures~\ref{fig:bert_xlm_vs_chlstm_by_family_by_tag} and
\ref{fig:bert_xlm_vs_chlstm_by_family_by_pos}) and neither is it uniform across
language families. MLMs are much better at Slavic tasks
(\figref{bert_xlm_vs_chlstm_slavic}). MLMs also struggle with some tasks
(\ref{ss:difftask}) but these tasks seem to be even more difficult for chLSTM.

Another proof of the effect of pre-training is the subpar performance of
random models (\ref{ssec:randinit}). We try two kinds of randomized models
(with and without randomized embedding) and both are much worse than their
pre-training counterparts regardless of which MLM we use. Perturbations do not
seem to affect the results when using randomized models, suggesting that the
random models mostly rely on word identities rather than some higher level
morphosyntactic knowledge (\figref{rand_clms}).

\paragraph{Left context plays a bigger role in morphosyntax than right context}
One of our main findings is that the left context plays a more important role
in morphosyntax than the right context and this seems to hold for most
languages. The relative importance of the left context is clearly observed in
both MLMs and in chLSTM (\tabref{shapley_summary}). Considering that all three
models are architecturally symmetrical, we can conclude that this is due to
linguistic reasons rather than some modeling bias. The simplest way we show
that the left context is more important is via the context masking
perturbations that mask words on the left, the right or both sides of the
target word. The effect of \pLtwo is clearly larger than \pRtwo
(c.f.~\tabref{perturbation_summary}) for all three models. \pLtwo is larger
than \pRtwo for all of the large language families
(\figref{context_masking_case_by_family}) in our dataset, although all context
masking has negligible effect on Uralic tasks. We further quantify the role of
each contextual word in \secref{shapley} and discuss some notable exceptions,
particularly from the Indic family.

\paragraph{\xlm's embedding and vocabulary are better suited for multilingual
morphology than \mbert's} Given that \xlm's subword vocabulary is twice as
large as \mbert's, we can expect more language-specific subwords especially
for low resource languages. We show three qualitative evidence that the
embedding of \xlm is actually better than the embedding of \mbert and we are
reasonably certain that this cannot be attributed to differences in the
pre-training corpora, since the gap shrinks in the higher layers.  First, we
find that the average probing accuracy at the embedding layer
(\ref{ssec:layer_pooling}) not only is higher for \xlm than \mbert but the
performance drop in comparison to using the weighted sum of all layers is
smaller for \xlm than \mbert. Second, there is very little improvement between
probing the embedding layer of \xlm and its first Transformer layer. This is
not the case for \mbert, where we see a larger improvement
(c.f.~\figref{layerwise_probing}). And third, the randomized models show that
not randomizing the embedding layer leads to superior performance by \xlm
compared to \mbert.

\paragraph{Trends pertain to specific morphosyntactic tags rather than
languages} Our extensive experiments show various trends and the languages
themselves are rarely the best unit for drawing general conclusions such
as one model being better than the other at a particular language. We find that
morphosyntactic tags and POS tags are both better choices as the units of
generalization. Whenever we group the results by language family
(Figures~\ref{fig:context_masking_case_by_family} and
\ref{fig:targ_permute_by_family}), the standard deviations are often larger
than the effects themselves.

We see examples where cross-POS comparison is meaningful, e.g. adjective tasks
rely on the context more than noun and verbal tasks do (in other words, \pTARG
has smaller effect on adjectives, see \ref{ssec:targ_permute}). But not all
POS categories lend themselves to similar generalizations, in particular
proper nouns are hard to make sense of as a cross-linguistically valid
grouping.\footnote{This is likely related to the fact that PROPN is the only
category where the typical member is a multi-word expression (MWE), and our
methods currently treat words that appear initial, middle, and final in an MWE
the same way, which distorts the ``true'' distance of a target from its
context.} We also show various individual examples where the tag, particularly
case, is a much better generalization horizon than the language itself
(Figures~\ref{fig:bert_xlm_vs_chlstm_by_family_by_tag} and
\ref{fig:context_masking_case_by_family}).

In \ref{ssec:shapley_generalization} we show quantitative proof for this claim
by computing the variance of the Shapley values as a function of the unit of
generalization. We find that language is a worse generalizing factor than both
tag and POS.

\section{Conclusions}

We introduced a dataset of \taskno probing tasks, covering \langno languages from \familyno families.
Using the dataset, we demonstrated that \mbert and \xlm embody considerable morphological
knowledge, reducing errors by a third compared to task-specific baselines with
the same number of trained parameters, and generally performing at a high level
even after masking portions of the input. 

Our main contribution is the detailed analysis of the role of the context by
means of perturbations to the data and Shapley values. We find that,
for this large suite of tasks,  the information resides dominantly in the
target word, and that left and right contexts are not symmetrical,
morphological processes are more forward spreading than backward spreading.

\section*{Acknowledgements} We are greatful for the anonymous reviewers'
insightful criticism, inquiries that prompted more discussion in the text, and
high-quality references. This work was partially supported by the the European Union
project RRF-2.3.1-21-2022-00004 within the framework of the Artificial
Intelligence National Laboratory Grant no RRF-2.3.1-21-2022-00004. Judit \'Acs
was partially supported by the Fulbright Scholarship and the Rosztoczy
Scholarship. Endre Hamerlik was partially supported by grant APVV-21-0114 of
the Slovak Research and Development Agency and Visegrad Scholarship nr.:
52110970. Competing interests: the authors declare none. 

\clearpage
\bibliography{nle_perturbation}
\bibliographystyle{nlelike}

\clearpage
\appendix

\section{Dataset statistics}\label{app:data_stats}

The dataset creation method and the choice of languages and tags are described
in \secref{data}. Here we list some additional statistics.

Our sampling method limits the sentence length between 3 and 40 tokens. The
average sentence length is 20.5 tokens. The subword tokenizer of \mbert
generates 38.2 tokens on average, while the subword tokenizer of \xlm outputs
34.2 tokens. Target fertility is defined as the number of subwords the target
token is split into. Target fertility is 3.1 and 2.6 for \mbert and \xlm
respectively. However, this measure varies substantially among languages,
particularly in the case of \mbert's tokenizer. \mbert generates the fewest
subwords for target words in English (2.05) and the most for Greek (4.75). \xlm
on the other hand tends to split Persian the least (2.05) with English coming
second (2.14) and it splits Hungarian the most (3.21) and the target fertility
for all other languages is below 3 for \xlm. \mbert's target fertility is above
3 for 22 out of the \langno languages. This again suggests that \xlm's larger
vocabulary is better suited for a multilingual setup.

Lastly we analyze the target words in more detail. Our data creation method
disallows having the same target word appear in more than one of the train,
validation or test splits. If the example sentence ``I \textit{read} your
letter yesterday.'' is part of the train set and \textit{read} is the target
word, it may not appear as a target word in the validation or the test set
regardless of its morphosyntactic analysis. It may be part of the rest of the
sentence though. A probing task has 2024 unique target words on average.
Split-wise this number is 1644 for the train split, 189 for the validation
split and 190 for the test split. Recall that we have 2000 train, 200
validation and 200 test sentences. Interestingly there is very little ambiguity
in the morphosyntactic analysis of the 4 tags we consider. 6.5\% of words have
ambiguous analysis in the full UD treebanks of the \langno languages.

\section{Stanza Setup}\label{app:stanza_setup}

Stanza is a collection of linguistic analysis tools in 70 languages trained on
UD treebanks. Stanza is an ideal tool for comparison since it was trained on
the same dataset we use to sample our probing tasks. The underlying model is a
highway bidirectional LSTM \citep{Srivastava:2015} with inputs coming from the
concatenation of word2vec or fastText vectors, an embedding trained on the
frequent words of UD and a character-level embedding generated by a
unidirectional LSTM. We use the default models available for each language.

It works in a pipeline fashion and one of its intermediate steps is
morphosyntactic analysis in UD format. We use this analysis as a high quality
baseline for all of our \langno languages and \taskno tasks except Albanian (6
tasks). We apply Stanza on the full probing sentences and we extract the
morphosyntactic analysis of the target words. Since Stanza's own tokenizer
often outputs a different tokenization than UD's gold tokenization, we extract
every overlapping token and check the union of the morphosyntactic tags if
there is more than one such overlapping token. This results in a set of
\verb|Feature=Value| pairs. We extract each occurrence of the feature of the
particular probing task. If there is only one and it is the same as the
reference label, it is correct. If there are more than one values and the
correct value is among them, we divide one by the number of distinct values. In
practice, this is very rare, 91\% of the time, there is only one value in it is
the same as the reference value. 8.3\% of the time the reference value is not
in the analysis of Stanza. The remaining 0.7\% are the cases where there are
multiple values including the \emph{correct} one.

\section{Layer Weights}\label{app:layer_weights}

\begin{figure}
    \centering
    \includegraphics[clip,trim={0 0 0 0cm},width=.99\textwidth]{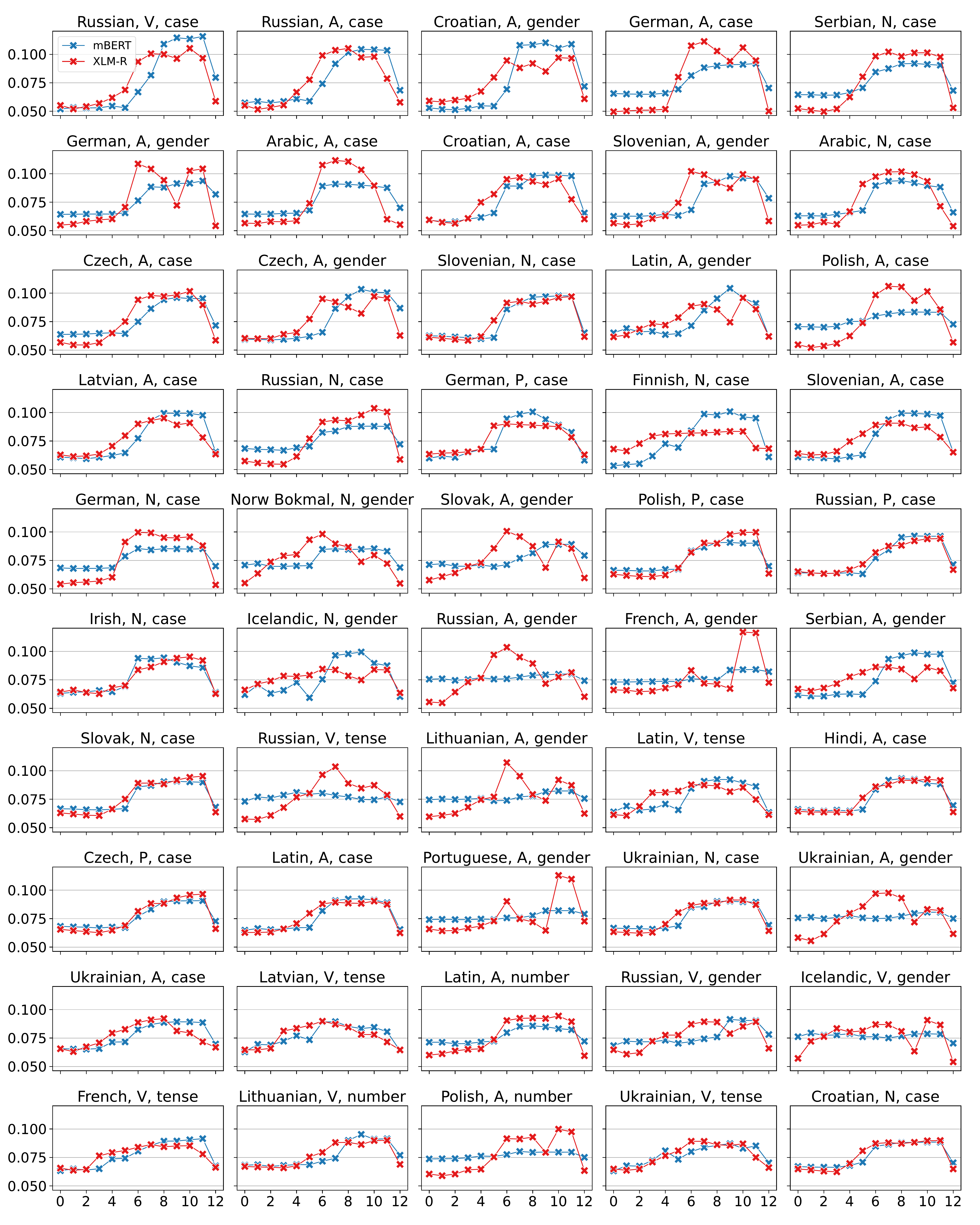}
    \vspace*{.2cm}
    \caption{Layer weight outliers. Layer 0 is the embedding layer.}
    \label{fig:layer_weight_outliers}
\end{figure}

We use scalar weighting for the 12 Transformer layers and the embedding layer.
The 13 scalar weights are learned for each probing experiment along with the
MLP weights. We use the softmax function to produce a probability distribution
for the layer weights. Our analysis shows that the layer distribution is very
close to the uniform distribution. We quantify this in two ways: the ratio of
the highest and the lowest layer weight, and the entropy of the weight
distribution. The two measures highly correlate (0.95 Pearson
correlation). The highest ratio of the largest and smallest layer weight is
2.22 for \mbert and 2.24 for \xlm, while the lowest entropy is 3.62 for both
models (the entropy of the uniform distribution is 3.7). We list the 50 task
with the highest max-min ratio in \figref{layer_weight_outliers}. Interestingly
the higher layers are weighted higher in each of the 50 outlier tasks somewhat
contradicting the notion that low level tasks such as morphology use lower
layers of BERT \citep{Tenney:2019a}.

We also found that the two models show similar patterns task-wise. When the layer
weights corresponding to one task have a lower entropy, they tend to have a
lower entropy when we probe the other model too. The ranking of the entropy of
the \taskno tasks' weights confirms this (0.63 Spearman's rank correlation).
We did not find any significant pattern pertaining to a particular language or
language family.

\section{Additional Shapley Figures}\label{app:additional_shapley}

We include extra figures and analysis for the Shapley value analysis from
\secref{shapley}.

\begin{figure}
    \centering
    \includegraphics[clip,trim={0 0 0
    0cm},width=.79\textwidth]{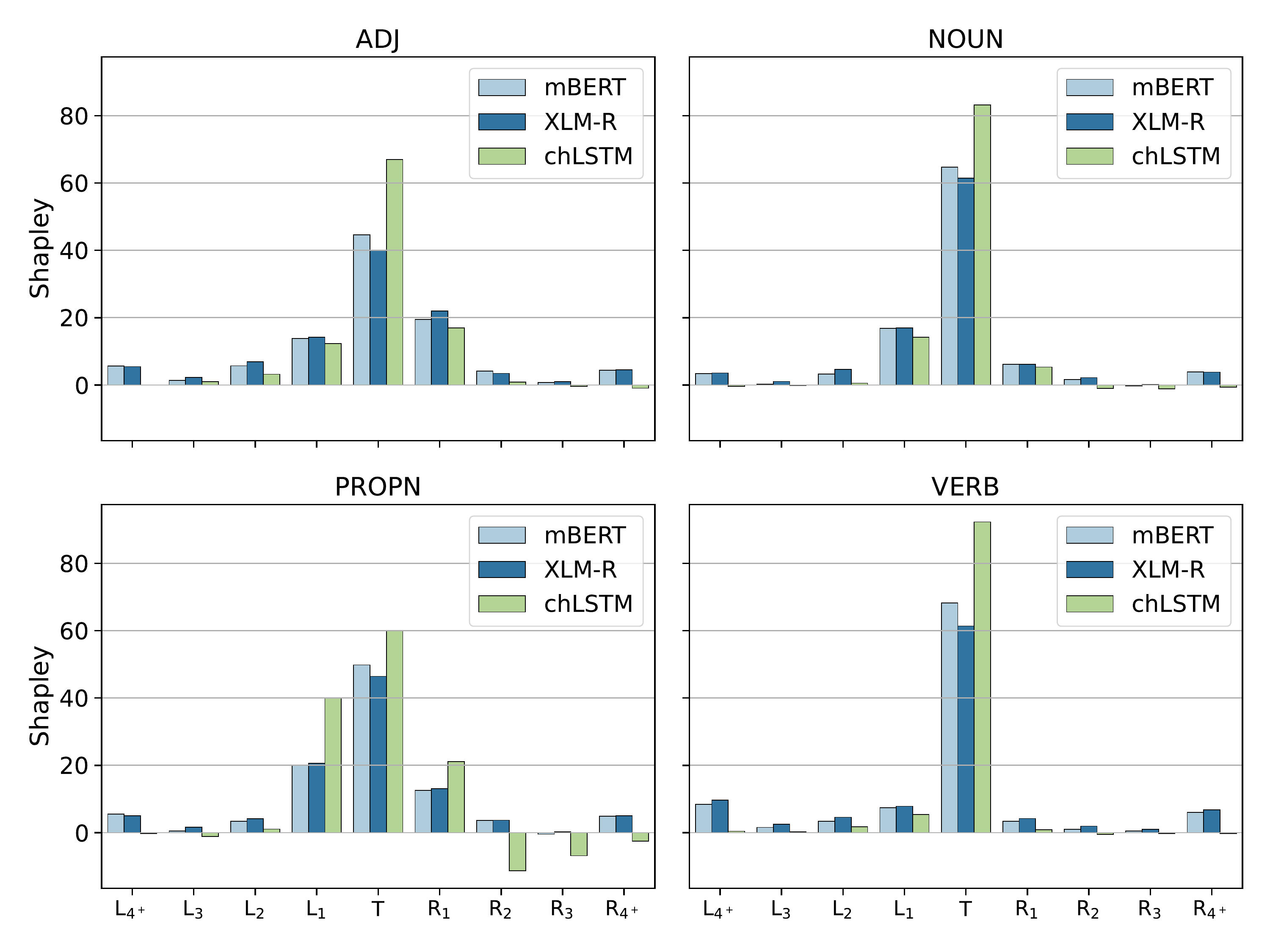}
    \vspace*{.2cm}
    \caption{Shapley values by POS and model.}
    \label{fig:shapley_per_pos}
\end{figure}

\begin{figure}
    \centering
    \includegraphics[clip,trim={0 0 0
    0cm},width=.79\textwidth]{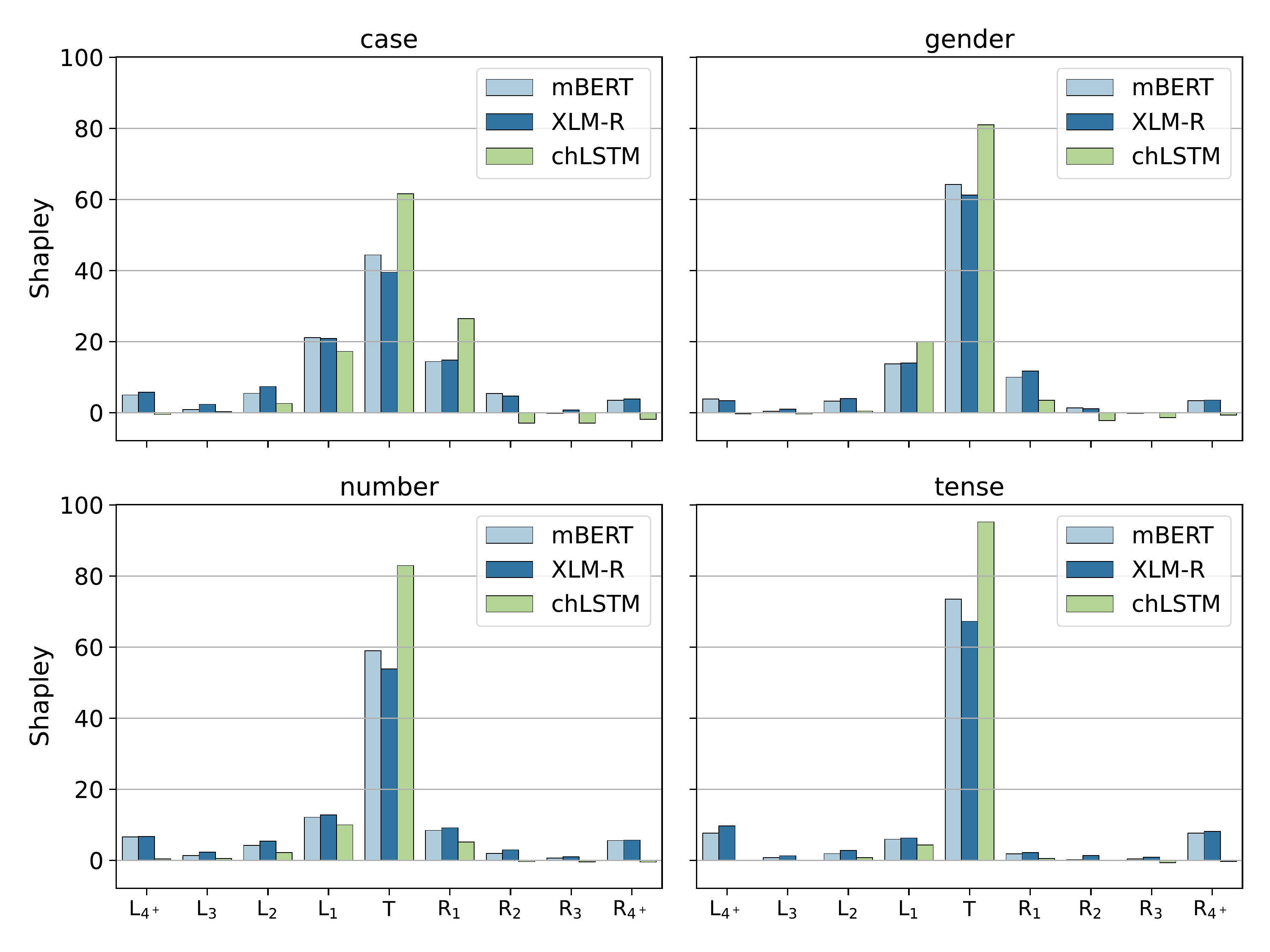}
    \vspace*{.2cm}
    \caption{Shapley values by POS and model.}
    \label{fig:shapley_per_tag}
\end{figure}

\figref{shapley_per_pos} shows the Shapley values by POS. Adjectives and proper
nouns rely on the context more than common nouns or verbs. We see similar
trends when we look at the per-tag Shapley values in
\figref{shapley_per_tag}. Tense, an exclusively verbal tag, has the largest
Shapley value for target out of the four tags.

\begin{figure}
    \centering
    \includegraphics[clip,trim={0 0 0
    0cm},width=.98\textwidth]{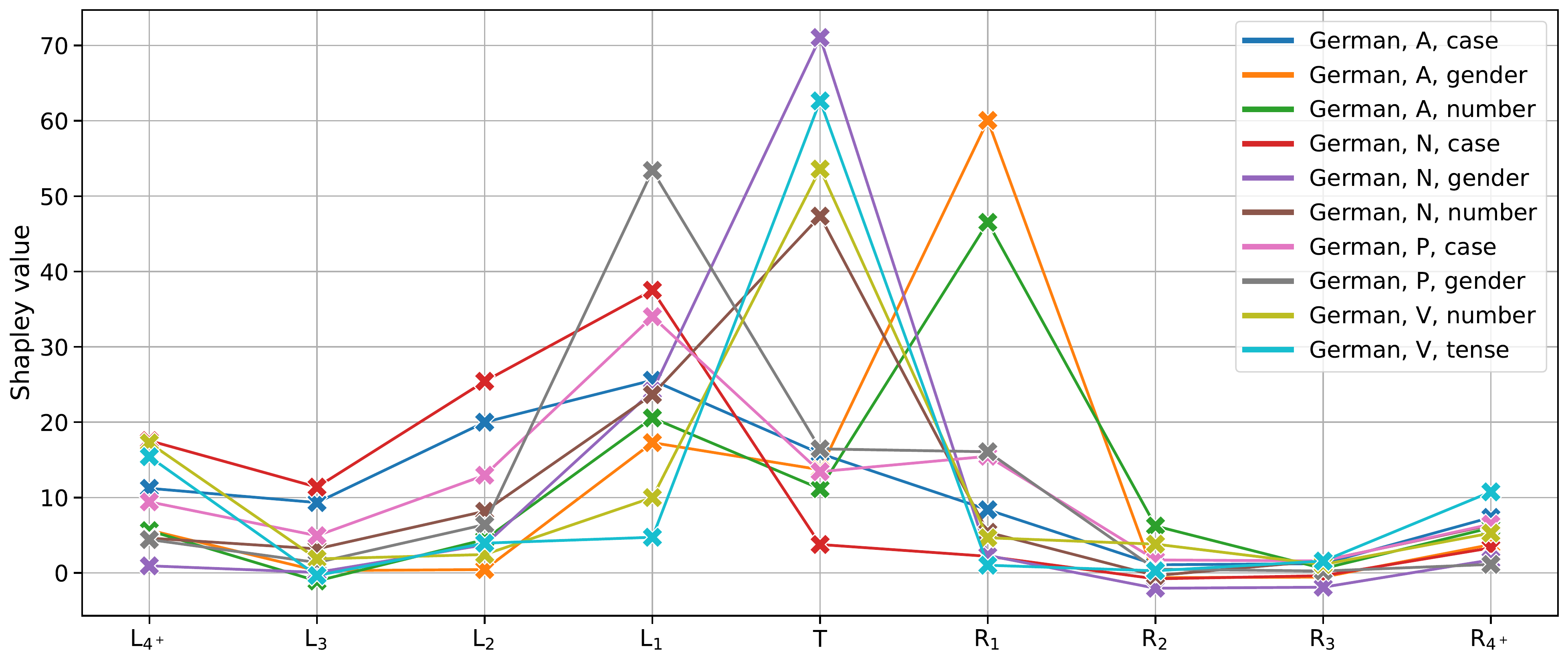}
    \vspace{0.3cm}
    \caption{Shapley values in German tasks.}
    \label{fig:shapley_german}
\end{figure}

Finally, we opted to mention the curious case of German tasks depicted in
\figref{shapley_german}. German is clearly the biggest outlier from the
Germanic family. In fact, it is dissimilar to every other language as
illustrated by our clustering experiments (c.f.~\figref{lang_clustering}).
Multiple German tasks appear in the main Shapley outliers
(\figref{shapley_outliers}) and among the tasks with the least uniform layer
weights (\figref{layer_weight_outliers}). When looking at the individual tasks'
Shapley values, only 4 of the 10 German tasks are dominated by the target. Case
tasks are left dominant, while adjective gender and number rely on $R_1$ the
most.

\section{Computational Requirements}

We run every unperturbed and perturbed experiment 10 times and report the
average of the 10 runs. The Shapley value computation requires $2^9=512$
experiments for each model and each task, we only run them once. Experiments
related to the ablations are also run once. The overall number of experiments
we ran is 460k. The average runtime of an experiment is 7 seconds and the total
runtime is roughly 38 days on a single GPU. We used a GeForce RTX 2080
Ti (12GB) and a Tesla V100 (16GB). The maximum number of epochs was set to 200
but in practice this is only reached in 2\% of the experiments. Early stopping
based on the development loss ends the experiments after 22 epochs on average.

\end{document}